# Deep learning for inverse problems with unknown operator[*]

## Miguel del Álamo

*University of Twente, Drienerlolaan 5, 7522 NB Enschede, The Netherlands*

*Institute for Mathematical Stochastics, University of Göttingen, Goldschmidtstr 7, 37077 Göttingen, Germany*
*e-mail:* miguel.del-alamo@mathematik.uni-goettingen.de

**Abstract:** We consider ill-posed inverse problems where the forward operator $T$ is unknown, and instead we have access to training data consisting of functions $f_i$ and their noisy images $Tf_i$. This is a practically relevant and challenging problem which current methods are able to solve only under strong assumptions on the training set. Here we propose a new method that requires minimal assumptions on the data, and prove reconstruction rates that depend on the number of training points and the noise level. We show that, in the regime of "many" training data, the method is minimax optimal. The proposed method employs a type of convolutional neural networks (U-nets) and empirical risk minimization in order to "fit" the unknown operator. In a nutshell, our approach is based on two ideas: the first is to relate U-nets to multiscale decompositions such as wavelets, thereby linking them to the existing theory, and the second is to use the hierarchical structure of U-nets and the low number of parameters of convolutional neural nets to prove entropy bounds that are practically useful. A significant difference with the existing works on neural networks in nonparametric statistics is that we use them to approximate operators and not functions, which we argue is mathematically more natural and technically more convenient.

## 1. Introduction

We consider a linear inverse problem, where our goal is to recover a function $f$ given observations from $Tf$ in a white noise regression model [47]

$$dY(x) = Tf(x)\,dx + \sigma\,dW(x), \quad x \in \mathbb{M} \tag{1}$$

where $\mathbb{M} \subset \mathbb{R}^d$ is a bounded set, $T$ is a linear, bounded operator in $L^2$ and $dW$ is a white noise process (see Section 2.1.2 in [19]). We assume that $f \in \mathcal{C}_L^s := \{f \in C^s \,|\, \operatorname{supp} f \subseteq [0,1]^d, \quad \|f\|_{C^s} \leq L\}$ is an $s$-Hölder function for $s, L > 0$.

---
[*]The author was funded by the Deutsche Forschungsgemeinschaft (DFG; German Research Foundation) Postdoctoral Fellowship AL 2483/1-1.

The noise level $\sigma > 0$ is assumed to be known, as it can otherwise be estimated efficiently [40, 45].

While linear inverse problems are well understood, in this paper we consider the challenging task of solving the inverse problem (1) when the operator $T$ is *unknown*. Instead, we have access to "training data", which consist of pairs of functions $f_i \in \mathcal{C}_L^s$ and their images

$$dY_i(x) = Tf_i(x)\,dx + \sigma\,dW_i(x), \quad x \in \mathbb{M}, \quad i = 1, \ldots, N \tag{2}$$

where $f_i$ are functions drawn from a probability distribution $\Pi$ in $\mathcal{C}_L^s$ (see Definition 2), and $dW_i$ are independent white noise processes defined in the same probability space.

Inverse problems with unknown operator appear in applications where the data generating mechanism is (partially) unknown or numerically too demanding to be used in practice. One example of that is physical systems where $T$ is not explicitly known but computed in intensive finite element simulations [48]. Another example are instrumental variable models in econometrics, where the unknown in the operator arises from the unknown correlations between the variables of interest [20].

Not knowing the operator makes the problem quite challenging. If the operator was known, solving the problem would mean roughly speaking to find a pseudoinverse of the operator $T$ (technically, of $T^*T$), and apply it to a denoised version of the data (1). Since the operator is unknown, we can only rely on the training data in order to construct an approximate pseudoinverse. In the past twenty years, several methods have been proposed for solving this problem under a quite strong assumption: the functions $f_i$ in the training set have to form an orthonormal system in $L^2$ (see e.g. [17, 37] and Subsection 1.1 below for a discussion). If the training data is so structured, it is easy to construct a pseudoinverse even explicitly, as done in [17]. However, it is unrealistic to have such structured training data in practice, and it is hence of interest to ask what can be done for more generic training data.

In this paper we consider that question by imposing a minimal assumption on the training data: the functions $f_i$ are drawn independently from a common probability distribution $\Pi$. This distribution represents the type of signals commonly appearing in a particular application, e.g. brain images in tomography.

Now, when dealing with such unstructured training data, it is not straightforward how to extract from it information about the operator $T$ and its pseudoinverse, as opposed to the explicit formula available under the orthonormality assumption. We approach this problem with the current *learning paradigm*, meaning that we look for a method that performs the inversion well in the given training data. Specifically, we employ neural networks as methods, and look for the neural net that minimizes the sum of squared residuals over the training data

$$\hat{F} \in \operatorname*{argmin}_{F \in \mathcal{F}} \sum_{i=1}^N \|f_i - F(Y_i)\|^2, \tag{3}$$

where $\mathcal{F}$ denotes a class of neural networks to be specified later. The intuition behind this method is as follows: if the network $\hat{F}$ is able to approximately solve the inverse problem for all the training data, then we expect it to approximately solve it for *similar* data, that is for data drawn from the same distribution $\Pi$.

One caveat at this point: we do not consider optimization issues in this work, and assume that we just have a neural network $\hat{F}$ that (somehow) minimizes the risk. However, we prove our results in a way that they apply for neural nets that *approximately* minimize the empirical risk in a precise sense (see Definition 6). This is a slight improvement, as it would be unrealistic to pretend to find the exact minimizer of the nonconvex minimization problem (3), see Remark 6.

Our focus in this paper is on the theoretical analysis of the (approximate) minimizer $\hat{F}$. In particular, we want to quantify its accuracy for solving the inverse problem (1), i.e., we want to show that $\|\hat{F}(Y) - f\|$ tends to zero in an appropriate sense as $N \to \infty$ and $\sigma \to 0$. Our analysis provides such a convergence in a quantitative way, with Theorem 1 proving that

$$\mathbb{E}_{f,f_i \sim \Pi, W, W_i} \|\hat{F}(Tf + \sigma \, dW) - f\|_{L^2}^2 \leq C \, L^2 \, \max\left\{\sigma^{\frac{4s}{2s+2\beta+d}}, N^{-\frac{s}{2s+2\beta+3d/2}}\right\} \quad (4)$$

up to logarithmic factors for $\beta \geq 0$, where the operator $T$ belongs to a class of $\beta$-smoothing linear operators that includes convolution operators, see Definition 1 and Example 1. This statement proves that the intuition of empirical risk minimization applies here: the network $\hat{F}$, trained to approximately invert the training data, is able to solve the inverse problem in unseen data drawn from the same distribution.

Furthermore, equation (4) shows that, if we have enough training samples ($N \geq \sigma^{-4\frac{2s+2\beta+3d/2}{2s+2\beta+d}}$), then the convergence rate is the minimax rate over $s$-Hölder functions for $\beta$-smoothing inverse problems (see Theorem 3 in [8]). This is remarkable, as it means that, if enough training data is given, our method performs roughly the same as if the operator was known. Pinpointing this behavior quantitative is of interest also from an epistemological viewpoint, as it represents the transition from classical model-based methods to modern model-agnostic training data-based methods.

We prove statement (4) for the approximate minimizer $\hat{F}$ over a class $\mathcal{F}$ of neural networks that have the architecture of U-nets, proposed in [42] for image segmentation tasks. These are convolutional neural networks with skip connections that have been used for inverse problems before, see Section 1.2 below. Crucially, our analysis gives us quantitative prescriptions for the choice of all parameters in the U-net (see Assumption 3 and equation (14)), thus providing a guide for the practitioner in the difficult task of choosing them.

The proof of our convergence result is based on two steps: first, we prove approximation properties for our class of U-nets. This is done by relating U-nets to the thresholding operator with respect to the wavelet-vaguelette system associated with the operator $T$. In a second step, we bound the complexity of the class of SU-nets in terms of its covering numbers. This is the most challenging step and the heart of the paper, and the central argument here is to consider the stability of SU-nets under perturbations of their coefficients. We refer to Section 2.4 for a detailed blueprint of the proof.

### *1.1. Inverse problems with unknown operator*

As remarked above, inverse problems with unknown operator pose a relevant and challenging problem. In the past 20 years, several approaches have been proposed to deal with them. The first work was that of Efromovich and Koltchiskii [17], which given training data builds a linear estimator for the pseudoinverse of $T^*T$. Their approach allows to have a different noise level in the training and the test data. Further, they recover the minimax rate of estimation if the training sample is large enough (specifically, if $N \geq \sigma^{-\frac{1}{2s+2\beta+1}}$ in dimension $d = 1$). A similar approach was proposed by Marteau [37], while a new type of nonlinear estimator with two variants was introduced by Hoffmann and Reiß [21]. They considered a (roughly equivalent) model where instead of having noisy training data, the operator is perturbed by a random, additive term. This corresponds to our setting having infinite training data and potentially different noise level in the training and the test data. Their estimators are shown to attain the minimax rates. A somewhat simplified model was considered by Cavalier and Hengartner [9], Johannes and Schwarz [26] and Marteu and Sapatinas [39], where the training data was supposed to consist of eigenfunctions of the operator $T$, whence the eigenvalues are observed with noise. In that setting, the different problem of estimating a quadratic functional was considered recently by Kroll [32].

A different estimator was proposed by Rosenbaum and Tsybakov [43] in the context of high-dimensional regression where the design matrix is observed with noise. This is a sort of discrete counterpart of the problem considered here. Their estimator is given by a Lasso-type method modified so that the noise in the design matrix is taken care of. Finally, a different approach was introduced by Trabs [46], who proposed a Bayesian method for the joint estimation of the unknown function $f$ and the operator $T$. His setting is similar to that in [21] in that he observes the operator $T$ up to additive random noise, instead of observing training data. His work is also novel in that he considers different degrees of uncertainty in the operator, which is parametrized by a unknown term that might be finite or infinite dimensional.

For completeness we mention here a family of works dealing with inverse problems with unknown operator that arise in instrumental variable models in econometrics. These works also correspond to the setting where the operator $T$ is known up to noise, because it is estimated from data in a certain way. The work of Hall and Horowitz [20] provides a good overview of this set of works until 2005, as well as two kernel-based nonparametric estimators with convergence rates. Further developments in this direction were given by Johannes et al [27], Marteau and Loubes [38], and Loubes and Pelletier [34], where they employ an interesting problem dependent form of regularization.

We remark that, in all the methods described above, one either effectively has infinitely many training data, or the training data $\{f_i\}$ is assumed to form an orthonormal system. This is at the heart of the constructions used, e.g. in [17], as the orthonormality is key for building their linear approximation to the pseudoinverse. We also remark that this assumption is quite restrictive and cannot be guaranteed in applications where the training data is either given, or has a very specific structure so that orthonormality cannot be naturally imposed (think e.g. of tomography images). In this sense, we find our approach more satisfying and realistic, as it does not impose any conditions on the training data besides being drawn independently from a probability distribution.

### *1.2. Why neural networks*

As stressed above, the lack of structure on the training data makes it unfeasible to construct an explicit pseudoinverse of $T$ directly. Instead, we resort to finding a method that minimizes (3), thus fitting the training data well. For this approach to work, the class $\mathcal{F}$ over which we minimize in (3) should be large enough to contain an operator that approximately inverts $T$. This requires a lot of expressive power, as the operator is unknown and can be anything within a class of smoothing operators (see Definition 1). In this work, we consider the case where $\mathcal{F}$ is a class of neural networks. Specifically, it will be a class of U-nets, proposed by [42]. Even though the U-net was originally proposed for the segmentation of MRI images, it has also been applied with minor modifications to proper inverse problems, such as limited angle X-ray tomography [24], 2-photon microscopy [33] and denoising of seismic data [36], to mention just a few applications. In this work, we will use a modification of U-nets, which we call simplified U-nets, or SU-nets, see Definition 5.

Let us give an intuition of why U-nets are suitable for our task. As argued in [42] and as can be seen in Figure 1, the U-net operates as follows: First, in the downward ("contracting") path, features are extracted from the input with the help of convolutional layers. The deeper we go in the vertical direction, the more features we extract, and the bigger the (effective) size of the convolutional kernels used. This means that the bottom layers carry information about lower frequencies, while higher frequency information is contained in the upper layers. Then, in the upward ("expanding") path, information from the deeper layers is combined with the layer immediately on top. This natural interpretation of U-nets in terms of frequencies is one of the key ideas of this work, as it will help us relate it to the wavelet transform, from what we will prove their approximation properties and their ability to invert an operator. See Definition 4 and Example 3 for the precise relation between these objects.

### *1.3. Literature on neural networks for nonparametric problems*

In the past few years, following the deployment and success of neural networks in practical applications, researchers in the mathematical statistics and inverse problems communities have begun to explain their performance in model problems with theories. As the literature is vast, in this section we will only mention the works directly relevant to us, that is those using neural nets for nonparametric problems such as regression and inverse problems and proving reconstruction rates. In that category, all of the relevant works study the approximation of functions with different types of neural networks under different assumptions. We refer to the classical works of Kolmogorov [31] for shallow networks, and Cybenko [10] and Hornik [22] for bounded activation functions. See the book by Anthony and Bartlett [2] for an overview of the developments until the 2000s. Since then, new techniques have allowed the treatment of deep networks and of unbounded activation functions, see e.g. [44] and [4] for results in a nonparametric statistics setting, and [7, 12, 16, 18, 30, 49, 50] and the review [13] for an approximation theoretic viewpoint. We also mention the works of D.-X. Zhou, specially [51], for the explicit treatment of convolutional networks.

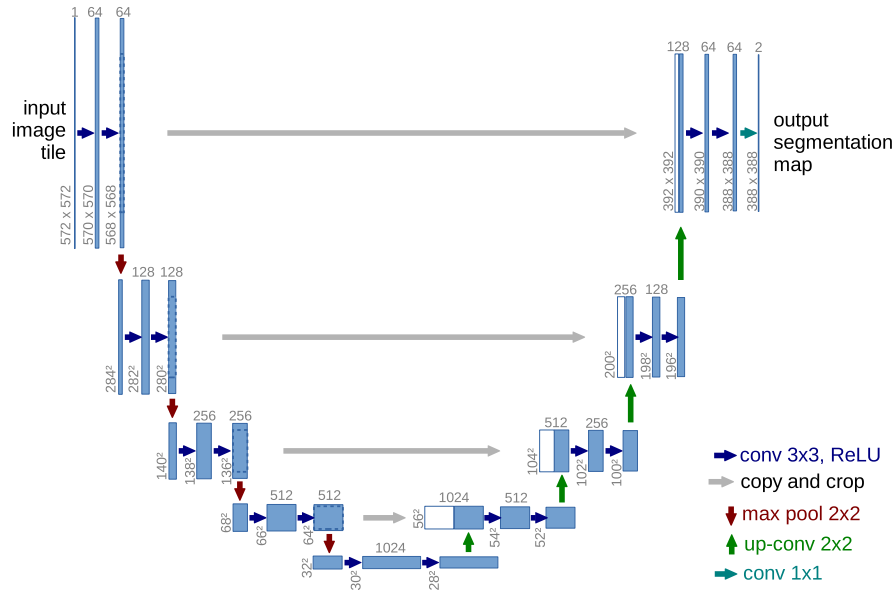

FIG 1. *U-net architecture, as proposed by [42] for segmentation of MRI images. Image taken from [42], reproduced with permission of Olaf Ronneberger.*

The results proven in these works have a common structure: they consider the approximation and entropy properties of a class of neural networks in terms of their complexity (measured by covering numbers and depending on the architecture, number of nonzero parameters, depth, width, etc), and their performance is given by a trade-off between these. In this respect, our strategy coincides with theirs. There is however a crucial difference between our use of neural nets and theirs: in all the papers mentioned above, the goal is to approximate a *function* given function values (that is, a regression problem or an inverse problem). In our case, the network approximates the *solution operator* of an inverse problem. This is a big difference, both in terms of the techniques used and of the implications.

In this sense, one of the contributions of this paper is to present a different way of modeling neural networks in statistical applications: as operators (or functionals) rather than as functions. We believe that this viewpoint is more natural in certain applications beyond the one we present here, such as segmentation and other image processing tasks (where the neural net is naturally used as an operator), or classification of complex objects such as image and sound (where the neural net is used as a functional). The naive answer to this statement might be that one can approximate functionals with functions, and operators with high-dimensional functions, and so use the function approximation properties of neural networks. Our answer to that has two parts: first, if we substitute an operator with a function from $\mathbb{R}^{n_1}$ to $\mathbb{R}^{n_2}$, the current theory gives an estimation error of the order $O(n_2)$ (times the noise level to a power $O(n_1^{-1})$). In particular the dependence on $n_2$ is terrible and likely to make the results useless. We take this as a hint that it is incorrect, as well as unnatural to model operators as high-dimensional functions in this setting. Of course, the answer to this is that one has to impose the right regularity properties. This brings us to our second argument, namely that in some applications such as image processing tasks, the object we look for naturally has *operator-like* properties, such as locality, boundedness, maybe even a representation as a Fourier integral operator, but it is not apparent, rather debatable, whether it has *function-like* properties, such as pointwise smoothness.

Finally, we want to mention that some researchers have already proposed to use neural networks to approximate the solution operator in inverse problems, see e.g. [1] and the very good review in [3]. Their motivation is however not to solve inverse problems with unknown operators, but to learn the right regularizer or regularization parameter for the problem. Furthermore, to the best of our knowledge these approaches do not prove reconstruction rates in terms of the smoothness, noise level and number of training data.

### *1.4. Contributions*

To summarize, in this work we make two contributions: First, we apply a type of convolutional neural network to linear inverse problems, and prove quantitative convergence rates. For that, we build a theoretical framework based on relating the neural network architecture to the wavelet transform, and we bound the complexity of the class of neural nets in terms of covering numbers. Furthermore, our theoretical analysis gives clear prescriptions for the choice of the many free parameters of the neural net, such as depth, size of the kernels, etc. Finally, we stress that the present framework is promising in that it can be generalized and applied to other settings, such as classification problems.

As a second contribution, we propose a method for solving inverse problems with unknown operators and prove convergence results for it. As explained in Section 1.1, this is a challenging problem for which solutions are known only under restrictive assumptions. In contrast, our assumptions are significantly weaker than those in the literature, while our results are of the same quality. In particular, our convergence rate depends on the size of the training set and the noise level, and the minimax rate over Hölder spaces is attained for a large enough training set.

### *1.5. Organization of the paper*

The paper is organized as follows. In Section 2.1 we present some of the basic definitions and assumptions, in Section 2.2 we introduce SU-nets and compare them at length with the wavelet transform, in Section 2.3 we state our main result and in Section 2.4 we give an overview of the proof. A discussion of our results is given in Section 3, while the main proofs are in Section 4.

## 2. Results

### *2.1. Assumptions*

**Definition 1.** For constants $\beta \geq 0$ and $C_\mathcal{T}, a_1, a_2 > 0$, the class $\mathcal{T}_{\beta, C_\mathcal{T}} = \mathcal{T}$ is defined as the set of all linear operators $T : \text{dom}(T) \subseteq L^2(\mathbb{R}^d) \to L^2(\mathbb{R}^d)$ such that

(i) The operator norm of $T$ is bounded by $C_\mathcal{T}$
(ii) The operator $T$ is $\beta$-smoothing, meaning that it maps between Sobolev spaces according to $a_1 \|g\|_{H^{t-\beta}} \leq \|T^*g\|_{H^t} \leq a_2 \|g\|_{H^{t-\beta}}$ for any $g \in \text{dom}(T^*)$ and any $t \geq 0$
(iii) The operator $T$ is translation invariant, meaning that it commutes with translations, i.e., if the translation operator is defined by $\tau_h g(x) := g(x+h)$, then $T\tau_h g = \tau_h Tg$ for all $g \in \text{dom}(T)$ and all $h \in \mathbb{R}^d$.

**Assumption 1** (Wavelet-vaguelette decomposition). Given a bounded set $\mathbb{M} \subset \mathbb{R}^d$, the linear operator $T$ and a father Daubechies wavelet $\varphi_{J,k,0}$ (at resolution level $J$), we assume that there are functions $\psi_k(x) \in L^2$ whose support is contained in the set $\mathbb{M}$ and such that for any function $g \in \text{dom}(T)$, we have

$$\langle Tg, \psi_k \rangle_{L^2} = \langle g, \varphi_{J,k,0} \rangle_{L^2}.$$

The set of functions $\psi_k$ is the so-called vaguelette system associated to $T$ and the Daubechies wavelet basis, as introduced in [14].

**Example 1.** An example of operators belonging to the class $\mathcal{T}_{\beta, C_\mathcal{T}}$ and satisfying Assumption 1 are convolution operators with kernel decaying fast enough. This includes relevant inverse problems, such as models for microscopy and astronomy [6], and also the solution operators to certain differential equations with the method of Green's function [5].

Let us consider this example in more detail. On one hand, we see that a convolution operator with a kernel $K$ is bounded with operator norm equal to $\|K\|_{L^1}$. Concerning the support properties in Assumption 1, suppose that the operator is given by convolution with a kernel $K$ satisfying

$$\mathcal{F}K(\xi) = (1 + |\xi|^2)^{-L},$$

which is a $2L$-smoothing operator, $L \in \mathbb{N}$, where $\mathcal{F}$ denotes the Fourier transform. The associated vaguelette can be computed explicitly [14] and has the form

$$\psi_k(x) = 2^{Jd/2} \mathcal{F}^{-1}\left[\frac{\mathcal{F}[\varphi_{0,0,0}]}{\mathcal{F}[K](-2^J \cdot)}\right](2^J x - k) = 2^{Jd/2}(1-\Delta)^L \varphi_{0,0,0}(2^J x - k),$$

where $\Delta$ denotes the Laplacian in $\mathbb{R}^d$. From this equation we read out that supp $\psi_k \subseteq$ supp $\varphi_{0,k,0}$, which is compact for Daubechies wavelets. In this case, the bounded set $\mathbb{M}$ in Assumption 1 can be taken to be the compact support of the Daubechies wavelets at scale $J = 0$.

**Remark 1.** The translation invariance assumption in Definition 1 is made for a particular reason: it guarantees that the vaguelettes $\psi_k(\cdot)$ in Assumption 1 are given by translations of a single function $\psi(\cdot)$, i.e., $\psi_k(x) = \psi(x - k \, 2^{-J})$. This will simplify the structure of the neural networks employed later. We remark, however, that our approach can be extended to operators that are not translation invariant at the price of a more complex network architecture. See Section 3 for a discussion.

**Definition 2** (Probability distribution in $\mathcal{C}^s$)**.** For $s > 0$, consider the set $\mathcal{C}^s$ of $s$-Hölder functions supported in the unit cube $[0,1]^d$, which is a metric space with the metric induced by the $s$-Hölder norm. Let $\mathcal{B}$ denote the Borel $\sigma$-algebra induced by the topology given by its norm. On the measurable space $(\mathcal{C}^s, \mathcal{B})$ we consider probability measures $\Pi$ defined with the usual axioms, see e.g. Chapter II in [41]. In this paper, we will only work with measures $\Pi$ supported in the bounded set $\mathcal{C}_L^s := \{f \in \mathcal{C}^s \,|\, \text{supp } f \subseteq [0,1]^d, \ \|f\|_{\mathcal{C}^s} \leq L\}$ for some $L > 0$.

**Example 2.** An example of a distribution as in Definition 2 is a Gaussian prior on the wavelet coefficients with an appropriate variance. More precisely, we consider random functions $f$ whose wavelet decomposition with respect to a suitable wavelet basis is given by

$$f = \sum_{j=0}^{J} \sum_{k=0}^{2^{jd}-1} \sum_{e} \alpha_{j,k,e} \varphi_{j,k,e} \text{ for } \alpha_{j,k,e} \sim \mathcal{N}(0, L^2 \, 2^{j(d-2s)}).$$

This ensures that $\mathbb{E}\|f\|_{\mathcal{C}^s} \leq L$. Further, the constraint in the support of the function $f$ can be achieved by using compactly supported wavelets (e.g. Daubechies) and by choosing the largest scale carefully.

### *2.2. Formalization of SU-nets*

In this section we introduced a class of neural networks which we call SU-nets. They are a modification of the U-net introduced in [42]. Before we define the nets in Definition 4, we motivate their structure by comparing them with the computation of the wavelet transform.

**Remark 2** (Wavelet transform via recursion)**.** In statistics, the wavelet coefficients of a function $g$ with respect to a wavelet basis $\{\varphi_{j,k,e}\}$ are usually computed and denoted by the inner products $c_{j,k,e} = \langle g, \varphi_{j,k,e}\rangle$, where $\varphi_{j,k,e}(x) = 2^{jd/2}\varphi_e(2^j x - k)$. In typical statistical applications one does not consider the computation of the coefficients any further. In computer science, however, the question of computation is in focus. The wavelet coefficients of a function can be computed in several ways, one of which is specially efficient and will be useful for our formulation of U-nets: the recursive computation. In a nutshell, we begin computing the wavelet coefficients of the signal at the smallest scale, and compute the ones at larger scales recursively.

In equations, we begin with the coefficients at the smallest scale, say $J$, *for the father wavelet only*: $c_{J,k,0} = \langle g, \varphi_{J,k,0}\rangle$. The rest of the coefficients are computed recursively as follows

$$c_{j-1,k,0} = \sum_l h[l - 2k] c_{j,l,0} = \sum_l h[l] c_{j,l+2k,0}$$
$$c_{j-1,k,e} = \sum_l g^e[l - 2k] c_{j,l,0} = \sum_l g^e[l] c_{j,l+2k,0}$$

for coefficients $h[l]$ and $g^e[l]$ given by

$$h[l] = 2^{-d/2}\int \varphi_0(t/2)\overline{\varphi_0(t-l)}\,dt, \quad g^e[l] = 2^{-d/2}\int \varphi_e(t/2)\overline{\varphi_0(t-l)}\,dt, \quad l \in \mathbb{Z}^d. \tag{5}$$

The coefficients also satisfy a recursion in the other direction, i.e. from big to small scales:

$$c_{j,k,0} = \sum_r h[r] c_{j-1,\frac{k-r}{2},0} + \sum_e g^e[r]\, c_{j-1,\frac{k-r}{2},e} \tag{6}$$

where the sum over $r$ runs over all indices $r \in \mathbb{Z}^d$ such that $(k-r)/2 \in \mathbb{Z}^d$. We refer to Theorem 7.10 in [35] for the proof of these statements. A warning here is due: Mallat uses the scale index in the *opposite* direction than we do, that is he writes $\phi_{j,k}(t) := 2^{-j/2}\phi(2^{-j}t - k)$. Beware of this change in notation when comparing the results.

While the computations above apply to all wavelets, in this paper we will use Daubechies wavelets [11], which are convenient due to their support and smoothness properties. Furthermore, using them gives some constraints in the filters $h[l]$ and $g^e[l]$, which we explain now. Let $\{\varphi_{j,k,e}\}$ be a family of one-dimensional Daubechies wavelets with $M \in \mathbb{N}$ vanishing moments. It follows from Theorem 4.2.10 in [19] that

$$\operatorname{supp}\varphi_{0,0,0} \subseteq [0, 2M - 1], \quad \operatorname{supp}\varphi_{0,0,1} \subseteq [-M + 1, M].$$

For a $d$-dimensional basis of Daubechies wavelets constructed by tensorization, this support property propagates in the obvious way. In particular, we see that for such a basis, the Lebesgue measure of the support of its elements is $|\text{supp } \varphi_{0,0,e}| = (2M-1)^d$. Furthermore, by the definitions of $h[l]$ and $g^e[l]$ in (5), we see that they also have a support of size

$$\#\text{supp } h = \#\text{supp } g^e \leq (2M-1)^d.$$

We can deduce further properties of the filters $h[l]$ and $g^e[l]$. By Corollary 4.2.4 in [19], using that $\int \varphi_{0,0,0}(x)\,dx = 1$ it follows that

$$\sum_l |h[l]|^2 = \sum_l |g^e[l]|^2 = 1.$$

More generally, these filters satisfy a certain orthogonality property, but we will not need it here.

**Definition 3** (Convolution operations on tensors). 1) For $d \in \mathbb{N}$, we say that $A$ is a *d-tensor* if it is a $d$-dimensional array of real coefficients.
2) Given a $d$-tensor, we define its *support size* or simply its *size* as its number of components. We will often work with $d$-tensors indexed by an equidistant regular grid of $2^{Jd}$ points in $[0,1]^d$, which we denote by $\Gamma_J$, for $J \in \mathbb{N}$.
3) Let $A$ and $\gamma$ be two $d$-tensors. We define their *downsampled convolution* as the $d$-tensor given by

$$\gamma *_\downarrow A(k) = \sum_l \gamma_l A_{2k-l},$$

where $k \in \mathbb{Z}^d$ and the sum runs over all indices $l$ such that $\gamma_l$ and $A_{2k-l}$ are defined.
4) Let $A$ and $\gamma$ be two $d$-tensors. We define their *upsampled convolution* as the $d$-tensor given by

$$\gamma *_\uparrow A(k) = \sum_l \gamma_l A_{\frac{k+l}{2}},$$

where $k \in \mathbb{Z}^d$ and the sum runs over all indices $l$ such that $(k+l)/2 \in \mathbb{Z}^d$ and $\gamma_l$ and $A_{(k+l)/2}$ are defined.

**Definition 4** (Structure of the SU-net). Let $d, J, S_{\text{filter}} \in \mathbb{N}$, and let $\psi, \phi \in L^2([0,1]^d)$ be continuous functions.

For $j = 0, \ldots, J-1$ and $e \in \{1, \ldots, 2^d - 1\}$, let $\alpha^{(j)}, a^{(j)}, \beta^{(j),e}, b^{(j),e}$ be $d$-tensors with support size $S_{\text{filter}}$, and let $\tau_j \geq 0$ be real numbers.

An operator $F(\cdot)$ that maps a function $g \in L^2([0,1]^d)$ to function $F(g)$ is called a *Simplified U-net*, or SU-net, if its output is given by (9) according to the following steps:

- First layer: given the input function $g \in L^2$, define the $d$-tensor

$$s_k^{(J)} = \int g(x)\psi(x - k\,2^{-J})\,dx \quad \text{for } k \in \Gamma_J,$$

with $\Gamma_J$ as in Definition 3.

- Contracting path:

$$s^{(j)} = \alpha^{(j)} *_\downarrow s^{(j+1)}$$
$$d^{(j),e} = \beta^{(j),e} *_\downarrow s^{(j+1)}, \quad \text{for } j = 0, \ldots, J-1, \quad \text{for } e = 1, \ldots, 2^d - 1$$

- Activation functions:

$$\overline{s}^{(0)} := s^{(0)}, \quad \overline{d}^{(j),e} = \rho(d^{(j),e}) := \max\{d^{(j),e} - \tau_j, 0\} - \max\{-d^{(j),e} - \tau_j, 0\} \tag{7}$$

where the nonlinearity acts coefficient-wise in the tensors.
- Expanding path:

$$\overline{s}^{(j)} = a^{(j-1)} *_\uparrow \overline{s}^{(j-1)} + \sum_e b^{(j-1),e} *_\uparrow \overline{d}^{(j-1),e} \quad \text{for } j = 1, \ldots, J \tag{8}$$

- Last layer:

$$F(g)(x) := \sum_k \overline{s}^{(J)}_k \phi(x - k\, 2^{-J}) \tag{9}$$

where the sum runs over all indices of the tensor $\overline{s}^{(J)}$, which are indexed by the grid $\Gamma_J$.

A graphical representation of the SU-net is given in Figure 2. Notice that the support of the tensors $s^{(l)}$ and $d^{(l),e}$ decreases as $l$ decreases: more precisely, it is divided by $2^d$ every time $l$ decreases by one. Conversely, the support size of $\overline{s}^{(l)}$ is multiplied by $2^d$ every time $l$ increases by one. This is akin to the pooling operation in U-nets, where it is usually performed either as max-pooling or average-pooling.

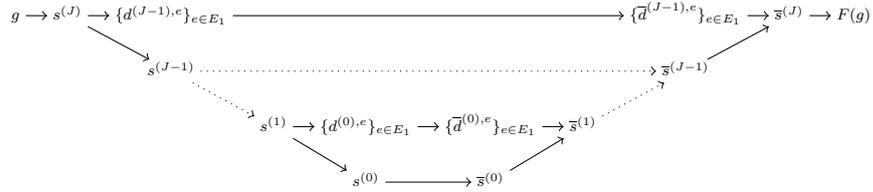

Fig 2. *Graphical representation of a Simplified U-net with depth $J$ in dimension $d$, where $E_1 = \{1, \ldots, 2^d - 1\}$.*

By definition, SU-nets are operators that map functions to functions. Their definition is such that the class of SU-nets contains the well-known operator that performs wavelet thresholding.

**Example 3** (Wavelet thresholding as an SU-net). Let $\varphi_{j,k,e}$ be a basis of Daubechies wavelets with $M$ vanishing moments, and let $h[l]$ and $g^e[l]$ be the corresponding filters as in Remark 2.

Consider the SU-net with filters given by

$$h[r] = \alpha^{(j)}_{-r} = a^{(j)}_{-r} \ \forall j, r$$
$$g^e[r] = \beta^{(j),e}_{-r} = b^{(j),e}_{-r} \ \forall j, r, e.$$

They hence have support $S_{\text{filter}} \leq (2M-1)^d$, and norm given by

$$\sum_r |\alpha^{(j)}_r|^2 = \sum_r |a^{(j)}_r|^2 = \sum_r |\beta^{(j),e}_r|^2 = \sum_r |b^{(j),e}_r|^2 = 1.$$

Let further the functions $\psi$ and $\phi$ in the SU-net be given by

$$\psi(x) = \phi(x) = 2^{Jd/2}\varphi_0(2^J x),$$

i.e. by the father wavelet at scale $J$. Consequently, we see that the wavelet coefficients of the signal $g$ are given by

$$c_{j,k,e} = d^{(j),e}_k, \ \ e \neq 0, \quad \text{and} \quad c_{j,k,0} = s^{(j)}_k,$$

that is the values of the $d$-tensors computed by the downward path of the SU-net. Notice further that the activation functions in the SU-net perform soft thresholding in the coefficients $c_{j,k,e} = d^{(j),e}_k$ with threshold $\tau_j$, thus yielding thresholded coefficients $\overline{c}_{j,k,e}$.

Finally, note that the upward ("expanding") path recomputes the wavelet coefficients that the smallest scale according to equation (6), and the last layer multiplies the thresholded coefficients with the father wavelet to yield the reconstructed function

$$F(g)(x) = \sum_k \overline{c}_{J,k,0} \, 2^{Jd/2}\varphi_0(2^J x - k),$$

which is the wavelet thresholded version of $g$.

We now further constraint the class of SU-nets that we will use by including some quantitative bounds.

**Definition 5** (Class of networks). Fix $J, r, R \in \mathbb{N}$, and let $\mathbb{M} \subset \mathbb{R}^d$ be the bounded set in Assumption 1. Given numbers $S_{\text{filter}} \in \mathbb{N}$ and $\kappa_\tau, C_{\psi,L^2}, C_{\psi,H^r} > 0$, the class $\mathcal{F}_J = \mathcal{F}_J(r, R, S_{\text{filter}}, \kappa_\tau, C_{\psi,L^2}, C_{\psi,H^r})$ consists of all SU-nets from Definition 4 with filters satisfying the following conditions:

1) The filters $\gamma \in \{\alpha^{(j)}, a^{(j)}, \beta^{(j),e}, b^{(j),e}, \ j = 0, \ldots, J-1, \ e = 1, \ldots, 2^d - 1\}$ are $d$-tensors satisfying

$$\#\text{supp } \gamma = S_{\text{filter}}, \quad \|\gamma\|_{\ell^2} \leq 1.$$

2) The thresholds satisfy $\tau_j \in [0, \kappa_\tau]$ for $j = 0 \ldots, J-1$.
3) The input filter $\psi$ belongs to the Sobolev space $H^r(\mathbb{M})$ and satisfies the following:

$$\|\psi\|_{L^2} \leq C_{\psi,L^2}, \quad \|\psi\|_{H^r} \leq C_{\psi,H^r}.$$

4) The output filter $\phi$ is *fixed* and given by $\phi(x) = 2^{Jd/2}\varphi(2^J x)$, where $\varphi(\cdot)$ is the father of a basis of Daubechies wavelets with $R$ continuous derivatives (and $6R+1$ vanishing moments and support size $(12R+1)^d$).

Definition 5 means that, in order to specify an SU-net from class $\mathcal{F}_J$, we need to choose the input filter $\psi$ (a continuous function), the compactly supported discrete filters $\alpha^{(j)}$, etc, and the thresholds $\tau_j$. The output filter in particular is fixed.

**Remark 3.** Following the construction and analogy in Example 3, we notice that the choice of the ReLU activation function in (7) is analogous to the soft-thresholding rule used in wavelet thresholding. Accordingly, different activation functions correspond to different thresholding rules (see page 202 in [28] for some examples), which then again have different motivations and interpretations. We wonder whether the statistical meaning of the tresholding rules can be transferred to the activation functions, thus providing guidance in their choice. We think for instance about the similarity between shrinkage in the James-Stein estimator and the batch normalization operation in neural networks.

**Remark 4** (Ensuring countability)**.** For theoretical purposes in Section 4.3, we will need the set $\mathcal{F}_J$ of SU-nets to the countable. In order to achieve that, we will define them as above, but all filters and thresholds taking real values will instead take rational values, and the continuous filter $\psi(x) \in H^r$ will be taken from a countable dense subset of the unit ball of $H^r$.

**Example 3** (Continued)**.** In order to accommodate the wavelet thresholding construction of Example 3 into the class $\mathcal{F}_J = \mathcal{F}_J(r, R, S_{\text{filter}}, \kappa_\tau, C_{\psi,L^2}, C_{\psi,H^r})$ of SU-nets, we need to choose a wavelet basis with $M = 1 + 6R$ vanishing moments, as it implies that the wavelet functions have $\lfloor 0.18 \cdot (M-1) \rfloor \geq R$ continuous derivatives and belong to $H^r$ for $R > r$ (see Remark 4.2.11 in [19]). This implies that we need to choose

$$S_{\text{filters}} = (12R+1)^d.$$

Further, we need to choose $C_{\psi,L^2} = 1$ and $C_{\psi,H^r} = \|\varphi_{0,0,0}\|_{H^r} 2^{Jr} = 2^{Jr}$ (see (10) for the last equality). With this choice of parameters, the wavelet thresholding operator as described in Example 3 belongs to $\mathcal{F}_J$.

We use this property in Section 4.2 below to analyze the approximation properties of SU-nets.

**Remark 5.** Notice that while the first layer in an SU-net is a convolutional layer, it is a *continuous* convolution, which is not standard in the deep learning community. We stress that this continuity is enforced for mathematical convenience, see the remarks after Assumption 2, and it would be discretized in practice.

Further, we remark that having the first layer as a convolution is enough for our purposes here, since we are dealing with translation invariant operators, for which the associated vaguelettes are all shifts of a single function, see Example 4 below.

For more general operators, we may need to perform more complex operations in the first layer in order to invert the operator locally. For non translation invariant operators, this could be done with the wavelet-vaguelette decomposition for general (not translational invariant) operators [14] or with a suitable Galerkin inversion as in [21]. In that case, one layer may not be enough to perform these operations, and several layers may be needed.

**Example 4** (Wavelet-Vaguelette transform). Let $T$ be an operator satisfying Definition 1 and Assumption 1. Consider the SU-net constructed in Example 3 above, with the only difference that $\varphi$ is a basis of Daubechies wavelets with $R = 1 + 6(r + \beta)$ vanishing moments, and the function $\psi$ is taken to be the vaguelette associated with the basis of wavelets and operator $T$. In this case, the corresponding SU-net performs thresholding with respect to the wavelet-vaguelette decomposition of the operator. This operation was proposed by [14] for solving statistical inverse problems.

In order for a class of SU-nets $\mathcal{F}_J$ as in Definition 5 to accommodate this operator, we need to choose the parameters such that

$$C_{\psi,L^2} = \frac{\|\varphi_{0,0,0}\|_{H^\beta}}{a_1} 2^{J\beta}, \quad C_{\psi,H^r} = \frac{\|\varphi_{0,0,0}\|_{H^{\beta+r}}}{a_1} 2^{J(\beta+r)}$$

for the constant $a_1 > 0$ in Definition 1. This is so because of the $\beta$-smoothing property of the operator $T$, which implies that $\|\psi\|_{L^2} \leq a_1^{-1} \|\varphi_{J,0,0}\|_{H^\beta} = a_1^{-1} 2^{J\beta} \|\varphi_{0,0,0}\|_{H^\beta}$ and $\|\psi\|_{H^r} \leq a_1^{-1} \|\varphi_{J,0,0}\|_{H^{\beta+r}} = a_1^{-1} 2^{J(\beta+r)} \|\varphi_{0,0,0}\|_{H^{\beta+r}}$. Moreover, the $H^t$-Sobolev norm of a sufficiently smooth father wavelet equals one for any smoothness degree $t$, as can be seen from the definition

$$\|\varphi_{0,0,0}\|_{H^t}^2 = \sum_{j,k,e} 2^{2jt} |\underbrace{\langle \varphi_{0,0,0}, \varphi_{j,k,e} \rangle}_{=\delta_{(j,k,e)=(0,0,0)}}|^2 = 1, \tag{10}$$

assuming that $\varphi_{j,k,e}$ is a family of Daubechies wavelets with at least $t$ continuous derivatives and more than $t$ vanishing moments.

**Example 5** (U-net). In the Simplified U-net, all filters are chosen freely from a bounded set, as in $\mathcal{F}_J$ in Definition 5. In particular, SU-nets are a particular subset of U-nets in [42] with the following properties:

- All layers are convolutional, with filters of size $S_{\text{filter}}$ in the middle layers.
- The activation functions are nonlinear in the skip connections, and linear otherwise,
- At each level, a constant number $2^d - 1$ of feature channels is used.

### *2.3. Main results*

**Assumption 2** (Training data). Let $\mathbb{M} \subset \mathbb{R}^d$ be a bounded set and $T \in \mathcal{T}_{\beta, C_T}$ be a $\beta$-smoothing operator as in Definition 1, both satisfying Assumption 1. For $s > 0$, let $C_L^s$ be the class of $s$-Hölder functions supported in $[0,1]^d$ with Hölder norm not larger than $L$. Further let $\Pi$ be a probability distribution on $C_L^s$, see Definition 2.

We call *training data* a set $\{(Y_i, f_i), i = 1, \ldots, N\}$ of observations given by $f_i \overset{i.i.d}{\sim} \Pi$ and

$$dY_i(x) = Tf(x)\,dx + \sigma\,dW_i(x), \quad x \in \mathbb{M}, \quad i = 1, \ldots, N,$$

where $dW_i$ are independent white noise processes defined in the same probability space.

**Notation.** In the following, we write $\mathbb{E}_f$ to represent the expectation with respect to the function $f$ drawn from the distribution $\Pi$. Accordingly, $\mathbb{E}_W$ represents the expectation with respect to the white noise process $W$.

Notice that we observe the transformed data $Y_i$ and the corresponding true signals $f_i$ as continuous data (i.e. at "full" resolution). This has several reasons: on one hand, we are analyzing SU-nets from the viewpoint of nonparametric statistics and inverse problems, so it is natural to estimate continuous functions, rather than discrete signals (which would fall in the realm of high-dimensional statistics). Finally, we choose to work with continuous signals for mathematical convenience: in particular, we leverage the approximation properties of wavelet bases, wavelet-vaguelette decompositions, etc, which are naturally expressed in the continuous setting. This being said, we remark that our results below can in principle be extended to discretely sampled $f_i$ and $dY_i$ in the training data, with the essential modifications being in Section 4.2, where we use the approximation properties of wavelets in order to prove approximation properties for SU-nets.

**Definition 6** (Approximately trained SU-net). Consider a set $\{(Y_i, f_i), i = 1, \ldots, N\}$ of training data as in Assumption 2, and a set $\mathcal{F}_J$ of SU-nets as in Definition 5.

1) We say that a *trained SU-net* $\hat{F}$ is a solution to the optimization problem

$$\hat{F}(\cdot) \in \underset{F \in \mathcal{F}_J}{\operatorname{argmin}} \sum_{i=1}^{N} \|F(Y_i) - f_i\|_{L^2}^2, \tag{11}$$

where we minimize over the filters $\psi \in H^r$, $\alpha^{(j)}, \beta^{(j),e}, a^{(j)}, b^{(j),e}$ and thresholds $\tau_j$.

2) For $\rho > 0$ fixed, we say that a network $\hat{F}(\cdot) \in \mathcal{F}_J$ is $\rho$-*approximately trained* if it satisfies

$$\sum_{i=1}^{N} \|\hat{F}(Y_i) - f_i\|_{L^2}^2 \leq \min_{F \in \mathcal{F}_J} \sum_{i=1}^{N} \|F(Y_i) - f_i\|_{L^2}^2 + \rho.$$

**Remark 6.** Clearly, a trained SU-net is also $\rho$-approximately trained for any $\rho > 0$. We consider $\rho$-approximately trained networks for an important reason: in practice, training is a difficult, nonconvex problem for which few guarantees are known (although first results look promising, see e.g. [25, 30] and references therein). In that sense, it is reasonable to allow deviations from the exact solution to the optimization problem (11), unattainable in practice, and instead consider approximate solutions which may be reached in practice. From the theoretical side, this relaxation is also valuable in order to find a approximate minimizer with rational weights, as required in Remark 4).

**Assumption 3** (Choice of parameters). Given $J, s, \beta, N, \sigma, a_1 > 0$, define the parameters

$$r = \max\{s, J, d/2 + 1\}, \quad R = r + \beta, \quad S_{\text{filter}} = (12\,R + 1)^d,$$
$$\kappa_\tau = \sigma\, 2^{J\beta} \log N, \quad C_{\psi,L^2} = \frac{2^{J\beta}}{a_1}, \quad C_{\psi,H^r} = \frac{2^{J(\beta+r)}}{a_1} \qquad (12)$$

and

$$\rho_{N,\sigma} := \max\left\{\sigma^{\frac{4s}{2s+2\beta+d}}, \sqrt{N}^{-\frac{2s}{2s+2\beta+3d/2}}(\log N)^3\right\}. \qquad (13)$$

**Remark 7.** The support size above corresponds to Daubechies wavelets with $R$ continuous derivatives and $M = 1 + 6 \cdot R$ vanishing moments, i.e. $(2M-1)^d$, see Example 3.

The choice of the smoothness follows from the need that Daubechies wavelets of smoothness $R$ approximate functions in $C^s$ well (which is needed in Proposition 2 below). Classical results (see e.g. Section 4.3.1 in [19]) imply that the wavelets need to have $M > s$ vanishing moments and $R > s$ continuous derivatives. In order to achieve this, we need to choose Daubechies wavelets with $M = 1 + 6 \cdot R$ vanishing moments and support $(12 \cdot R + 1)^d$ (see the continuation of Example 3 above).

Furthermore, for our bounds on the covering numbers of $\mathcal{F}_J$ in Proposition 3 we will need the smoothness of the wavelets to be as large as possible in order to reduce the size of our parameter space. There we are confronted with another effect: the larger the smoothness, the larger the support of the wavelets. And as stated in Proposition 3, if the support size $S_{\text{filter}} = (2M - 1)^d$ grows unboundedly, our entropy bound increases.

A reasonable compromise between these effects is to choose wavelets with $r = \max\{s, J\}$ continuous derivatives, where $J$ is the depth of our SU-net (see the proof of Proposition 4)

**Theorem 1.** Fix $s, \beta, \sigma, N, L, a_1 > 0$ and let

$$J = \left\lceil \frac{1}{2s + 2\beta + d} \log_2 \min\{\sigma^{-2}, \sqrt{N}^{\frac{2s+2\beta+d}{2s+2\beta+3d/2}}\} \right\rceil \qquad (14)$$

For $r, S_{\text{filter}}, C_{\psi,L^2}, C_{\psi,H^r}$ and $\rho_{N,\sigma}$ as in Assumption 3, let

$\hat{F} \in \mathcal{F}_J(r, R, S_{\text{filter}}, \kappa_\tau, C_{\psi,L^2}, C_{\psi,H^r})$ be a $\rho_{N,\sigma}$-approximately trained network as in Definition 6 with $N$ training data sampled from the probability distribution $\Pi$ from Definition 2 in $\mathcal{C}_L^s$ and with noise level $\sigma$. Assume that the data arises from a $\beta$-smoothing operator $T \in \mathcal{T}_{\beta, C_\mathcal{T}}$ as in Definition 1 and satisfying Assumption 1. Then there is a constant $C > 0$ independent of $\sigma, L$ and

$N$ such that

$$\sup_{T \in \mathcal{T}_{\beta,C_\mathcal{T}}} \mathbb{E}_{f,f_i \sim \Pi, W, W_i} \|\hat{F}(Tf + \sigma \, dW) - f\|_{L^2}^2 \tag{15}$$
$$\leq C\, C_\mathcal{T}^2\, L^2 \, \max\left\{ \sigma^{\frac{4s}{2s+2\beta+d}}, N^{-\frac{s}{2s+2\beta+3d/2}} (\log N)^3 \right\}.$$

*Proof.* The result follows from Proposition 4 below. □

We remark that the constant in (15) is explicit, and can be found in the proof of Proposition 4.

**Remark 8.** We stress that the expectation in (15) is taken with respect to the observed data $f$ following the distribution $\Pi$ (and also with respect to the training data $f_i$ coming from $\Pi$, and with respect to the noise in the training data $W_i$ and in the observed data $W$). On one hand, this means a clear limitation to the predictive power of the trained network. On the hand, it is only natural that a network trained on a specific set of data generalizes well on similar data, but not necessarily well on unrelated data. We find this result valuable in that it shows the limitations of neural networks trained on a limited set of data.

**Remark 9.** In the setting of Theorem 1, let the size of the training set be given by $\sqrt{N} = \sigma^{-2\gamma}$ for a parameter $\gamma > 0$. Then we have the following regimes:

- If $\gamma < \frac{2s+2\beta+3d/2}{2s+2\beta+d}$, the right-hand side of (15) behaves like $N^{-\frac{s}{2s+2\beta+3d/2}} (\log N)^3$, which is slower than the classical nonparametric estimation rate. This is the regime where we do not have enough training data, and the lack of knowledge on $T$ dominates the error. Still, the error tends to zero when $N$ increases.
- If $\gamma \geq \frac{2s+2\beta+3d/2}{2s+2\beta+d}$, then the right-hand side of (15) behaves like $\sigma^{\frac{4s}{2s+2\beta+d}}$ with an additional logarithm if equality holds. Ignoring the logarithm, this is the minimax rate of estimation for functions in $C_L^s$. This means that $\hat{F}$ performs optimally in this regime. In particular, here we have so many samples $N$ in the training set, that not knowing the operator is not an issue.

Regarding the optimality of the result in Theorem 1, we have seen that the right-hand side matches the minimax lower bound for $N$ large enough. On the other hand, if $N \leq \sigma^{-4\frac{2s+2\beta+3d/2}{2s+2\beta+d}}$, the rate is driven by $N$ and we do not know whether there is a matching lower bound in this regime. We suspect that the rate is in fact *suboptimal* in this regime, as the exponent of $N$ seems arbitrary and it arises from a complicated balance of terms in the entropy bound. As explained in Remark 11 below, one possibility could be to find an alternative way of approximating the vaguelettes, maybe using additional information on the operator. Generally, we suspect that a different approach to bound the entropy term could yield an improvement in the exponent, but we do not pursue this issue any further in this work.

*2.4. Structure of the proof*

With $\mathcal{F}_J = \mathcal{F}_J(r, R, S_{\text{filter}}, \kappa_\tau, C_{\psi,L^2}, C_{\psi,H^r})$, define the set of functionals

$$\mathcal{H}_J(r, R, S_{\text{filter}}, \kappa_\tau, C_{\psi,L^2}, C_{\psi,H^r}) = \left\{ h_F(g) := \mathbb{E}_W \|F(Tg + \sigma\, dW) - g\|_{L^2}^2 \,\bigg|\, F \in \mathcal{F}_J \right\},$$

that map functions $g \in \text{dom}(T)$ to nonnegative numbers. The expectation $\mathbb{E}_W$ is taken with respect to the white noise process $dW$.

In Proposition 1 we split the error into a stochastic error, an approximation error and the optimization error $\rho_{N,\sigma}$. The first two errors are taken care of in Propositions 2 and 3.

**Proposition 1.** Given training data $\{(Y_i, f_i), i = 1, \ldots, N\}$ as in Assumption 2 and let $\hat{F}$ be a $\rho_{N,\sigma}$-approximately trained SU-net as in Definition 6. Then we have

$$\mathbb{E}_{f,W,f_i,W_i} \|\hat{F}(Tf + \sigma\, dW) - f\|_{L^2}^2 \leq 2\, \mathbb{E}_{f_i} \sup_{h \in \mathcal{H}_J} |\mathbb{E}_f h(f) - N^{-1} \sum_{i=1}^N h(f_i)|$$
$$+ \inf_{F^* \in \mathcal{F}_J} \mathbb{E}_{f,W} \|F^*(Tf + \sigma\, dW) - f\|_{L^2}^2 + \rho_{N,\sigma}.$$

See Section 4.1 for the proof.

**Proposition 2** (Approximation error). For $\beta \geq 0$, let $T$ be a $\beta$-smoothing operator as in Definition 1 and satisfying Assumption 1. For $s, J \in \mathbb{N}$ and $r, S_{\text{filter}}, \kappa_\tau, C_{\psi,L^2}, C_{\psi,H^r}$ as in (12), consider the set of SU-nets
$\mathcal{F}_J = \mathcal{F}_J(r, R, S_{\text{filter}}, \kappa_\tau, C_{\psi,L^2}, C_{\psi,H^r})$. Then we have

$$\inf_{F \in \mathcal{F}_J} \mathbb{E}_{f,W} \|F(Y) - f\|_{L^2}^2 \leq C \left[ 2^{dJ + 2\beta J} \sigma^2 + L^2\, 2^{-2Js} \right] \tag{16}$$

for a constant $C > 0$ independent of $\sigma$ and $L$.

The proof is based on choosing $F$ to be the operator that performs thresholding with respect to the wavelet-vaguelette decomposition of $T$, see Section 4.2.

**Proposition 3** (Entropy bound). For $J, s, r, S_{\text{filter}} \in \mathbb{N}, r > d/2$, and constants $\kappa_\tau, C_{\psi,L^2}, C_{\psi,H^r} > 0$, consider the class $\mathcal{F}_J = \mathcal{F}_J(r, R, S_{\text{filter}}, \kappa_\tau, C_{\psi,L^2}, C_{\psi,H^r})$. Then we have

$$\mathbb{E}_{f_i} \sup_{F \in \mathcal{F}_J} \left| \frac{1}{N} \sum_{i=1}^N h_F(f_i) - \mathbb{E}_f h_F(f) \right|$$
$$\leq \overline{C} \frac{C_{\psi,L^2}^2\, 2^{Jd}\, J^3}{\sqrt{N}}\, \max\{S_{\text{filter}}, (\text{diam}(\mathbb{M}))^d\, C_{\psi,H^r})^{d/(2r)}\}$$

where

$$\overline{C} = 2L^2 + 8\sqrt{2} C_B \left( \sqrt{\frac{\log 2 + H_J}{J^2}} + 2\sqrt{2^{d+1} \frac{\log(52\, J)}{J}} + 2^{3(d+1)/2} \frac{52^{d/(2r)}}{1 - d/(2r)} \right)$$

and $C_B = 18 \, 2^{2d} \left( L^2 C_{\mathcal{T}}^2 + \sigma^2 \right)$ and $H_J = 2^d \, J \, S_{\text{filter}} \log(2 \max\{1, \kappa_\tau \, 2^{Jd/2}\})$, where $\|f_i\| \leq L$ and $\|T\|_{op} \leq C_{\mathcal{T}}$.

The proof of Proposition 3 is based on the usual metric entropy techniques and is given in Section 4.3.

**Remark 10.** In order to prove entropy bounds on $\mathcal{H}_J$, we apply the classical entropy integral bound to it. Since it is usually applied to classes of functions, and not to classes of *functionals*, we review here the key assumptions needed on $\mathcal{H}_J$.

First, let $(C_L^s, \mathcal{B})$ be a subset of $C^s$ endowed with the Borel $\sigma$-algebra $\mathcal{B}$ induced by the topology generated by the $s$-Hölder norm. From this viewpoint, $\mathcal{H}_J$ is a set of functions from the measurable space $(C_L^s, \mathcal{B})$ to the nonnegative real numbers (i.e. a set of functionals). Further, each functional $h \in \mathcal{H}_J$ is a measurable function from $(C_L^s, \mathcal{B})$ to the nonnegative reals endowed with the Borel $\sigma$-algebra. This is so, because $h(f)$ is a continuous function of $f$, since all operations done are linear, quadratic or Lipschitz (the activation functions).

Finally, in order to apply the entropy bound in Theorem 3.5.1 in [19] to $\mathcal{H}_J$, we need to show that it is countable. But this follows from the fact that the set of SU-nets is countable, see Remark 4.

All in all, we have that $(\mathcal{H}_J, \tilde{\rho}_J)$ is a separable pseudo-metric space with the pseudo-metric given by

$$\tilde{\rho}_N(h_{F_1}, h_{F_2}) := \sqrt{N^{-1} \sum_{i=1}^{N} |h_{F_1}(f_i) - h_{F_2}(f_i)|^2}, \tag{17}$$

where $f_i \sim \Pi$ are independent random functions from the training data.

**Remark 11.** The term $C_{\psi, H^r}^{d/(2r)}$ in the upper bound in Proposition 3 is responsible for the ackward-looking exponent $\frac{s}{2s+2\beta+3d/2}$ in the final convergence rate (15). The structure of the term $C_{\psi, H^r}^{d/(2r)}$ is intuitive from the viewpoint of empirical processes: we are dealing with the entropy of a set of $d$-dimensional and (roughly) $r$-differentiable functions, so bounds of the form $K^{d/2r}$ are expected, where $K$ is the radius of the set. The issue is that we need the upper bound $C_{\psi, H^r}$ to grow like $2^{J(r+\beta)}$ (see Assumption 3) in order for our network to be able to approximate vaguelettes of the right regularity. A possible improvement of this issue, and hence of the convergence rate in (15), would be to devise an alternative way of approximating the vaguelettes.

Now we are ready to prove our main result combining Propositions 1 to 3.

**Proposition 4.** Let $r, R, S_{\text{filter}}, C_{\psi, L^2}, C_{\psi, H^r}$ and $\rho_{N, \sigma}$ be as in Assumption 3. Let $\hat{F}$ be a $\rho_{N,\sigma}$-approximately trained SU-net as in Theorem 1. Assuming that $J$ is given by (14) we have the bound

$$\mathbb{E}_{f, W, f_i, W_i} \|\hat{F}_J(Y) - f\|_{L^2}^2 \leq C \, C_{\mathcal{T}}^2 \, L^2$$
$$\times \max \left\{ \sigma^{\frac{4s}{2s+2\beta+d}} + \frac{\sigma^{-2\frac{2\beta+3d/2}{2s+2\beta+d}}}{\sqrt{N}} (\log_2 \sigma^{-2})^3, \right.$$

$$\sqrt{N}^{-\frac{2s}{2s+2\beta+3d/2}}(\log\sqrt{N})^3 + \sigma^2 \sqrt{N}^{\frac{2\beta+d}{2s+2\beta+3d/2}}\Big\}$$

where the constant $C > 0$ is independent of $T, f, \sigma, L$ and $N$. In particular, the claim is uniform over operators $T \in \mathcal{T}_{\beta, C_\mathcal{T}}$ satisfying Definition 1 and Assumption 1.

*Proof.* First, choosing the parameters as in Assumption 3, the claim of Proposition 3 reads

$$\mathbb{E}_{f_i} \sup_{F \in \mathcal{F}_J} \left| \frac{1}{N} \sum_{i=1}^N h_F(f_i) - \mathbb{E}_f h_F(f) \right| \leq \underline{C} \, C_\mathcal{T}^2 L^2 \frac{2^{J(2\beta+3d/2)} J^3}{\sqrt{N}}$$

for

$$\underline{C} = \frac{\overline{C}}{C_\mathcal{T}^2 L^2} \, a_1^{-2-d/2r} \, 2^{\beta d/2} \, \mathrm{diam}(\mathbb{M})^{d^2/2r},$$

which is bounded by a constant independent of $J, \sigma, L, C_\mathcal{T}$ and $N$.

Combining Propositions 1 to 2 with this result we get

$$\sup_{T \in \mathcal{T}} \mathbb{E}_{f, W, f_i, W_i} \|\hat{F}_J(Y) - f\|_{L^2}^2$$

$$\leq 2\underline{C} \, C_\mathcal{T}^2 L^2 \frac{2^{J(2\beta+3d/2)} J^3}{\sqrt{N}} + C^2 L^2 2^{-2Js} + C^2 \, 2^{J(d+2\beta)} \, \sigma^2 + \rho_{N,\sigma}$$

$$\leq C \, C_\mathcal{T}^2 L^2 \left( \frac{2^{J(2\beta+3d/2)} J^3}{\sqrt{N}} + 2^{-2Js} + 2^{J(d+2\beta)} \, \sigma^2 \right) + \rho_{N,\sigma}. \tag{18}$$

Depending on the relation between $\sigma^2$ and $\sqrt{N}$, there are two different values of $J$ that minimize the above expression. In order to analyze that, we introduce the exponent $\gamma > 0$ defined such that $\sqrt{N} = \sigma^{-2\gamma}$.

**Oversampled regime:** $\gamma \geq \frac{2s+2\beta+3d/2}{2s+2\beta+d}$ In this case, the optimal choice of $J$ is $J = \frac{1}{2s+2\beta+d} \log_2 \sigma^{-2}$, which yields the bound

$$\mathbb{E}_{f,W,f_i,W_i}\|\hat{F}_J(Y) - f\|_{L^2}^2 \leq C \, C_\mathcal{T}^2 L^2 \left[\sigma^{\frac{4s}{2s+2\beta+d}} + \frac{\sigma^{-2\frac{2\beta+3d/2}{2s+2\beta+d}}}{\sqrt{N}}(\log_2 \sigma^{-2})^3\right] + \rho_{N,\sigma}.$$

Notice that the term $\rho_{N,\sigma}$ defined in (13) is of the same asymptotic order as the first two terms.

**Undersampled regime:** $\gamma < \frac{2s+2\beta+3d/2}{2s+2\beta+d}$

In this case, the optimal choice of $J$ is $J = \frac{1}{2s+2\beta+3d/2} \log_2(\sqrt{N})$, which yields the bound

$$\mathbb{E}_{f,W,f_i,W_i}\|\hat{F}_J(Y) - f\|_{L^2}^2$$
$$\leq C \, C_\mathcal{T}^2 L^2 \left[ \sqrt{N}^{-\frac{2s}{2s+2\beta+3d/2}} (\log \sqrt{N})^3 + \sigma^2 \sqrt{N}^{\frac{2\beta+d}{2s+2\beta+3d/2}} \right] + \rho_{N,\sigma}.$$

Here, again, the term $\rho_{N,\sigma}$ is of the same asymptotic order as the first term.

Finally, notice that in both regimes, the first terms dominate (i.e. are bigger), and only for the choice $\gamma = \frac{2s+2\beta+3d/2}{2s+2\beta+d}$ are both terms of the same polynomial order. □

## 3. Conclusion and outlook

In summary, in this paper we have considered inverse problems with unknown operators and proposed a method to solve them given training data. We proved a convergence rate for the method that depends on the noise level and the amount of training data, and that coincides with the minimax rate in the regime where the amount of data is above a threshold. We remark once again that the main advantage of our result over the available literature is that our assumption on the training data is minimal and realistic, while the assumptions used in the literature of inverse problems with unknown operators are strong and not always realistic.

Further, we have employed neural networks to solve our problem, hence showing how they can be applied to solve inverse problems in general. In particular, this is to the best of our knowledge the first work in which a convergence guaranty is proven for neural networks in inverse problems. Crucially, our analysis yields concrete prescriptions on how to choose the many hyperparameters of the network, such as its depth, the number of channels, the size of the convolution kernels, etc.

As it stands, our result concerns continuous neural network that are not employed in practice. As remarked above, this is done for mathematical convenience, in order to profit from the approximation theory available in the continuous world. We see this issue pragmatically, and merely think of the continuous neural net as a mathematical model for the discrete one.

Finally, we remark that in this paper we have proposed a different way of thinking about neural networks, that is, to see them as operators (or functionals), and have presented some techniques to analyze them. We see this however as a basic result proven with basic techniques, and both of then can be extended in a variety of directions, a few of which we mention here.

First, the generalization of the present results to more general operators, either lacking translation invariance or even a wavelet-vaguelette decomposition, is of interest. We suspect that dealing with this would require the architecture of the network, probably by introducing a block of dense layers before the SU-net kicks in.

Another generalization of interest is to allow a different noise level in the train and test data. This could be done again by modifying the network architecture and including a "module" that is able to estimate the noise level of the input, plus skip connections that bring the estimated noise level to the thresholds in the ReLU nonlinearities. For this extension, it is crucial to compute how this change in architecture affects the entropy estimates.

Further, it would be of interest to look into the distribution $\Pi$ in more depth:

on one hand, it would help to see how it is in practical applications (see e.g. the works of Mumford, such as [23]) and which regularity/structural properties it satisfies. Moreover, it is possible to adapt our results to the distribution $\Pi$ more precisely, and try to find the "right" regularity exponent $s$ and thus the optimal convergence rate for data from a distribution $\Pi$, and the corresponding lower bound. As a first approximation, one could consider the covering numbers of the support of $\Pi$, on not of the whole of $\mathcal{C}_L^s$, in the computation of the entropy bound.

As a last remark, we believe that our techniques can be used to analyze neural networks in other problems, such as segmentation or classification of high-dimensional (continuous) signals, for which they are not yet fully understood.

## 4. Proofs

In this section we prove the main propositions of the paper. For compactness of the notation, we employ a special notation for the expected values: $\mathbb{E}_f$ denotes the expected value with respect to the random function $f \sim \Pi$, drawn from the distribution $\Pi$. And $\mathbb{E}_W$ denotes the expected value with respect to the white noise process $dW$.

### *4.1. Proof of Proposition 1*

*Proof of Proposition 1.* Conditionally on the $f_i$'s, we have

$$\mathbb{E}_{f,W,W_1,\ldots W_N} \|\hat{F}(Tf + \sigma dW) - f\|_{L^2}^2$$
$$= \mathbb{E}_{f,W,W_1,\ldots W_N} \|\hat{F}(Tf + \sigma dW) - f\|_{L^2}^2$$
$$- \mathbb{E}_{W_1,\ldots W_N} \frac{1}{N} \sum_{j=1}^N \|\hat{F}(Tf_j + \sigma dW_j) - f_j\|_{L^2}^2$$
$$+ N^{-1} \mathbb{E}_{W_1,\ldots W_N} \sum_{j=1}^N \|\hat{F}(Tf_j + \sigma dW_j) - f_j\|_{L^2}^2. \tag{19}$$

The difference of the first two terms can be bounded as follows

$$\mathbb{E}_{f,W,W_1,\ldots W_N} \|\hat{F}(Tf + \sigma dW) - f\|_{L^2}^2 - \mathbb{E}_{W_1,\ldots W_N} \frac{1}{N} \sum_{j=1}^N \|\hat{F}(Tf_j + \sigma dW_j) - f_j\|_{L^2}^2$$

$$\leq \left| \mathbb{E}_{f,W} \mathbb{E}_{W_1,\ldots W_N} \|\hat{F}(Tf + \sigma dW) - f\|_{L^2}^2 \right.$$
$$\left. - \frac{1}{N} \sum_{j=1}^N \mathbb{E}_{W_1,\ldots W_N} \|\hat{F}(Tf_j + \sigma dW_j) - f_j\|_{L^2}^2 \right|$$

$$\leq \sup_{F \in \mathcal{F}_J} \left| \mathbb{E}_{f,W} \mathbb{E}_{W_1,\ldots W_N} \|F(Tf + \sigma dW) \right.$$

$$- f\|_{L^2}^2 - \frac{1}{N}\sum_{j=1}^{N}\mathbb{E}_{W_1,\ldots W_N}\|F(Tf_j + \sigma dW_j) - f_j\|_{L^2}^2\bigg|$$

$$\leq \sup_{F\in\mathcal{F}_J}\bigg|\mathbb{E}_{f,W}\|F(Tf+\sigma dW) - f\|_{L^2}^2 - \frac{1}{N}\sum_{j=1}^{N}\mathbb{E}_{W_j}\|F(Tf_j+\sigma dW_j) - f_j\|_{L^2}^2\bigg|$$

$$= \sup_{F\in\mathcal{F}_J}\bigg|\mathbb{E}_f h_F(f) - \frac{1}{N}\sum_{j=1}^{N} h_F(f_j)\bigg|,$$

where in the last equality we define the functions

$$h_F(f) = \mathbb{E}_W \|F(Tf + \sigma\, dW) - f\|_{L^2}^2$$

for $h_F \in \mathcal{H}_J$. The second term in (19) can be dealt with as follows: conditionally on $dW_1, \ldots, dW_N$ we have

$$\frac{1}{N}\sum_{j=1}^{N}\|\hat{F}(Tf_j + \sigma\, dW_j) - f_j\|_{L^2}^2 \leq \frac{1}{N}\sum_{j=1}^{N}\|F^*(Tf_j + \sigma\, dW_j) - f_j\|_{L^2}^2 + \rho_{N,\sigma}$$

for any $F^* \in \mathcal{F}_M$, using the properties of $\hat{F}$ in Definition 6. Consequently, using conditional expectations we get

$$\frac{1}{N}\mathbb{E}_{W_1,\ldots W_N}\sum_{j=1}^{N}\|\hat{F}(Tf_j + \sigma\, dW_j) - f_j\|_{L^2}^2$$

$$\leq \frac{1}{N}\mathbb{E}_{W_1,\ldots W_N}\sum_{j=1}^{N}\|F^*(Tf_j + \sigma\, dW_j) - f_j\|_{L^2}^2 + \rho_{N,\sigma}$$

$$= \frac{1}{N}\sum_{j=1}^{N}\underbrace{\mathbb{E}_{W_j}\|F^*(Tf_j + \sigma\, dW_j) - f_j\|_{L^2}^2}_{=h_{F^*}(f_j)} + \rho_{N,\sigma}$$

$$\leq \sup_{h\in\mathcal{H}_M}\bigg|\mathbb{E}_f h(f) - N^{-1}\sum_{j=1}^{N} h(f_j)\bigg| + \mathbb{E}_f\mathbb{E}_W\|F^*(Tf + \sigma\, dW) - f\|_{L^2}^2 + \rho_{N,\sigma},$$

and since the element $F^* \in \mathcal{F}_M$ was arbitrary, we can take the infimum in the right-hand side, which yields the claimed bound. □

### 4.2. Approximation properties

Here we prove Proposition 2 considering two separate cases for the sake of illustration: first, the case that $T$ is the identity operator, so we just have a denoising problem. In this case, we use an SU-net to perform wavelet thresholding, see Proposition 5. And second, the case where $T$ is a nontrivial smoothing operator, in which case we use an SU-net to perform thresholding with respect to its wavelet-vaguelette decomposition, see Proposition 6.

**Proposition 5** (Approximation in the regression model $(T = id)$)**.** Let $J \in \mathbb{N}$ be arbitrary and $L > 0$. Let $\Pi$ be a probability distribution on $\mathcal{C}_L^s$ as in Definition 2. Consider the class $\mathcal{F}_J = \mathcal{F}_J(r, R, S_{\text{filter}}, \kappa_\tau, C_{\psi, L^2}, C_{\psi, H^r})$ from Definition 5 with parameters as in Assumption 3. Then there is a network $F \in \mathcal{F}_J$ such that, for data

$$dY(x) = f(x)\, dx + \sigma\, dW(x),\ x \in [0,1]^d, \quad f \sim \Pi \tag{20}$$

with $dW$ and $f$ independent, $F(\cdot)$ achieves the estimation error

$$\mathbb{E}_{f \sim \Pi}\, \mathbb{E}_W \|F(Y) - f\|_{L^2} \leq C \left[ 2^{dJ/2}\, \sigma + L\, 2^{-Js} \right] \tag{21}$$

where $C > 0$ is a numerical constant independent of $J, \sigma$ and $L$.

The proof consists of two stages: first, we show that an SU-net can be used to compute the wavelet thresholding operator of a signal; and second, we apply the well known theory on wavelet thresholding to bound its accuracy.

*Proof.* Let $\varphi_{j,k,e}$ denote a family of Daubechies wavelets with $M = 1 + 6R$ vanishing moments, and hence $R > s$ continuous derivatives. Recall from Remark 2 that its filters $h$ and $g^e$ satisfy

$$\#\text{supp } h, \#\text{supp } g^e \leq (2M-1)^d = (12R+1)^d$$

and $\sum_m |h[m]|^2 = \sum_m |g^e[m]|^2 = 1$.

Let now $F$ be an SU-net in $\mathcal{F}_J$ as in Definition 5, with the discrete filters

$$\begin{aligned} a_m^{(j)} &= \alpha_m^{(j)} = h[m], \quad \forall j = 0, \ldots, J-1 \\ b_m^{(j),e} &= \beta_m^{(j),e} = g^e[m], \quad \forall j = 0, \ldots, J-1, \ \forall e = 1, \ldots, 2^d - 1, \end{aligned} \tag{22}$$

and the continuous filters chosen to be

$$\psi(x) = \phi(x) = 2^{Jd/2} \varphi_{0,0,0}(2^J x).$$

It is immediately verified that the filters satisfy the conditions of Definition 5 with the constants of Assumption 3. In particular we have

$$\begin{aligned} \|\psi\|_{L^2} &= \|\varphi_{0,0,0}\|_{L^2} = 1 \\ \|\psi\|_{H^r} &= 2^{Jr} \|\varphi_{0,0,0}\|_{H^r} = 2^{Jr}, \end{aligned}$$

since $\varphi_{0,0,0}$ has $R = r$ continuous derivatives, see equation (10) above. All in all, we conclude that the chosen SU-net $F$ indeed belongs to $\mathcal{F}_J$ with the chosen parameters.

Now, by construction the function $F(g)(x)$ is the wavelet thresholded version of $g$, where its wavelet coefficients $c_{j,k,e} = \langle g, \varphi_{j,k,e} \rangle$ are soft thresholded by a value $\tau_j > 0$. Classical results in nonparametric regression (see e.g. [15]) imply that choosing the threshold of the order $\tau_j = \sigma \sqrt{2j}$ yields a bound

$$\mathbb{E}_W \|F(Y) - f\|_{L^2} \leq c_1 2^{dJ/2} \sigma + c_2 \|f\|_{C^s} 2^{-Js} \tag{23}$$

where $\|f\|_{C^s}$ denotes the $s$-Hölder norm of $f$, and $c_1, c_2 > 0$ are numerical constants independent of $f$, $J$ and $\sigma$. Taking the supremum over $f \in \mathcal{C}_L^s$ gives the bound

$$\sup_{f \in \mathcal{C}_L^s} \mathbb{E}_W \|F(Y) - f\|_{L^2} \leq c_1 2^{dJ/2} \sigma + c_2 L 2^{-Js},$$

which we use to bound the expectation with respect to $f \sim \Pi$ using the inequality

$$\mathbb{E}_{f \sim \Pi} \mathbb{E}_W \|F(Y) - f\|_{L^2} \leq \sup_{f \in \mathcal{C}_L^s} \mathbb{E}_W \|F(Y) - f\|_{L^2}.$$

This completes the proof. □

**Proposition 6** (Approximation for inverse problems model ($T \neq id$)). Let $J \in \mathbb{N}$ be arbitrary. Consider the class of networks $\mathcal{F}_J = \mathcal{F}_J(r, R, S_{\text{filter}}, \kappa_\tau, C_{\psi,L^2}, C_{\psi,H^r})$ with parameters as in Assumption 3. Let the operator $T$ and the set $\mathbb{M}$ satisfy Definition 1 and Assumption 1. Then there is a network $F \in \mathcal{F}_J$ such that, given data

$$dY(x) = Tf(x) \, dx + \sigma \, dW(x), \ x \in \mathbb{M}, \quad \text{and } f \sim \Pi \text{ independently of } dW,$$

$F(\cdot)$ achieves the estimation error

$$\mathbb{E}_{f \sim \Pi} \mathbb{E}_W \|F(Y) - f\|_{L^2} \leq C \left[ 2^{dJ/2 + \beta J} \sigma + L 2^{-Js} \right]$$

where $\Pi$ is a probability distribution in $C_L^s$ and $C > 0$ is a numerical constant independent of $\sigma$, $J$ and $L$.

*Proof.* The proof essentially follows the same steps as the proof of Proposition 5, so we will just point out the differences. The difference is that here we do not want to do wavelet thresholding with an SU-net, but instead we want to threshold the wavelet vaguelette decomposition of the data. In order to do that, consider a basis of Daubechies wavelets $\varphi_{j,k,e}$ with $1 + 6R$ vanishing moments, just as in the proof of Proposition 5. We define hence an SU-net $F$ with discrete filters equal to the discrete filters of the Daubechies wavelets as in (22), and the continuous filter in the last layer is chosen to be a father wavelet, i.e., $\phi(x) = 2^{Jd/2} \varphi_{0,0,0}(2^J x)$.

For the continuous filter $\psi$ in the first layer of $F$, we consider the wavelet-vaguelette decomposition of the operator $T$, which exists by Assumption 1. Moreover, by the assumption of translation invariance in Definition 1, the vaguelettes are all given by translations of one function, which we denote by $\psi$. Due to the $\beta$-smoothing property of $T$, this function $\psi$, which is implicitly defined by the equation $T^*\psi = 2^{Jd/2} \varphi_{0,0,0}(2^J x)$, satisfies

$$\|\psi\|_{H^t} \leq a_1^{-1} \|T^*\psi\|_{H^{t+\beta}} \leq a_1^{-1} 2^{J(t+\beta)} \|\varphi_{0,0,0}\|_{H^{t+\beta}} \quad \text{for any } t \geq 0,$$

whence $\|\psi\|_{L^2} \leq a_1^{-1} 2^{J\beta} \|\varphi_{0,0,0}\|_{H^\beta} = a_1^{-1} 2^{J\beta}$ and $\|\psi\|_{H^r} \leq a_1^{-1} 2^{J(r+\beta)} \|\varphi_{0,0,0}\|_{H^{r+\beta}} \leq a_1^{-1} 2^{J(r+\beta)}$, using equation (10) to compute the Sobolev norms, and the fact that $\varphi_{0,0,0}$ has $R = r + \beta$ continuous derivatives. This choice of $\psi$ and

its bounds are hence in accordance with the parameters in Assumption 3, and we conclude that the constructed network $F$ belongs to $\mathcal{F}_J$ with the desired parameters.

But for this choice of the network, the first layer performs the following operation:

$$\langle Tf + \sigma dW, \psi(\cdot - k/2^{-J})\rangle$$
$$= \langle f, \varphi_{J,k,0}\rangle + \sigma \underbrace{\langle dW, \psi(\cdot - k\,2^{-J})\rangle}_{\sim \mathcal{N}(0, \|\psi(\cdot - k\,2^{-J})\|_{L^2}^2)} = \langle f, \varphi_{J,k,0}\rangle + \sigma \|\psi\|_{L^2}\, \widetilde{\epsilon}_k$$

for $\tilde{\epsilon}_k \sim N(0,1)$ no longer independent. This is however not an issue, as it is well-known that thresholding these observations with a slightly modified threshold yields optimal results (see e.g. [29]), where the modified threshold is of the same order of magnitude as the universal threshold $\sigma\sqrt{2J}$, so it is in our interval $[0, \sigma\, 2^{J\beta} \log N]$ of allowed thresholds.

With this choice of $\psi$ in the first layer, we have transformed the data into the coefficients with respect to the Daubechies wavelet basis, at the cost of blowing up the noise level by a factor $\|\psi\|_{L^2} = O(2^{J\beta})$ which is why we need to choose $C_{\psi,L^2} = O(2^{J\beta})$ and $\kappa_\tau = \sigma 2^{J\beta} \log N$.

The following layers in the SU-net perform the same operations as in Proposition 5: computation of the wavelet coefficients, thresholding, and synthesizing the function back together. The same theory as for the reconstruction properties of wavelets applies. Hence, taking the increased noise level into account, the final result (23) reads now

$$\mathbb{E}_W \|F(Y) - f\|_{L^2} \leq C_1 2^{dJ/2}\, 2^{J\beta}\, \sigma + C_2\, \|f\|_{C^s}\, 2^{-Js}.$$

The rest of the argument is now identical to the steps after equation (23) above, so we do not repeat it. □

### 4.3. Entropy bounds

In this section we prove Proposition 3 by applying the usual covering number techniques to the class $\mathcal{H}_J$ of functionals. The covering numbers of this class is then related to those of the class $\mathcal{F}_J$ of neural nets, which we bound in Section 4.3.2. The key ingredient for bounding the covering numbers is an analysis of the stability of neural networks under perturbations of their parameters, given in Section 4.4.

*Proof of Proposition 3.* We use Theorem 3.5.1 in [19] applied to the class of functions $\mathcal{H}_J = \mathcal{H}_J(r, R, S_{\text{filter}}, \kappa_\tau, C_{\psi,L^2}, C_{\psi,H^r})$, which maps $C_L^s$ functions to $\mathbb{R}$. $P_N$ is hence the empirical distribution of the training data $f_i \sim \Pi$ in $C_L^s$, and we have

$$\sqrt{\|P_N h^2\|_{\mathcal{H}_J}}$$
$$= \sup_{F \in \mathcal{F}_J} \sqrt{N^{-1} \sum_{i=1}^{N} |h_F(f_i)|^2} \leq 2L^2 + 2^{Jd} 4 C_{\psi,L^2}^2 (J 2^d + 1)^2 \left(L^2 C_{\mathcal{T}}^2 + \sigma^2\right) =: B \tag{24}$$

using Claim 2 below, as well as $\|f_i\| \leq L$ and $\|T\|_{op} \leq C_{\mathcal{T}}$. In this setting, applying Theorem 3.5.1 in [19] with the modification that $0 \notin \mathcal{H}_J$, which gives an additional term $T_0$, yields

$$\mathbb{E}_{f_i} \sup_{F \in \mathcal{F}_J} \left| \frac{1}{N} \sum_{i=1}^{N} h_F(f_i) - \mathbb{E}_f h_F(f) \right|$$
$$\leq 2 \frac{T_0}{\sqrt{N}} + \frac{8\sqrt{2}}{\sqrt{N}} \mathbb{E}_{f_i} \int_0^{\sqrt{\|P_N h^2\|_{\mathcal{H}_J}}} \sqrt{\log 2 D(\mathcal{H}_J, \tilde{\rho}_N, \tau)} \, d\tau$$

where $\tilde{\rho}_N(h_1, h_2)$ is the random semi-metric in (17) and $D(\mathcal{H}_J, \tilde{\rho}_N, \delta)$ denotes the $\delta$-packing numbers of $\mathcal{H}_J$ with respect to $\tilde{\rho}_N$. The additional term $T_0$ is given by

$$T_0 = \mathbb{E}_{f_i} \mathbb{E}_{\xi_i} \left| \frac{1}{\sqrt{N}} \sum_{i=1}^{N} \xi_i h_0(f_i) \right|$$

for an arbitrary element $h_0 \in \mathcal{H}_J$ and independent centered sub-Gaussian random variables $\xi_i$ with unit variance. We bound this term choosing the functional $h_0$ that corresponds to the *zero* SU-net $F(\cdot) = 0$, i.e. the network that always returns zero. This network belongs to $\mathcal{F}_J$, as we can take all the filters to be zero. For this choice of $h_0$, $T_0$ can be bounded as

$$T_0 = \mathbb{E}_{f_i} \mathbb{E}_{\xi_i} \left| \frac{1}{\sqrt{N}} \sum_{i=1}^{N} \xi_i \|f_i\|_{L^2}^2 \right|$$
$$\leq \sqrt{\mathbb{E}_{\xi_i} \left| \frac{1}{\sqrt{N}} \sum_{i=1}^{N} \xi_i \|f_i\|_{L^2}^2 \right|^2} = \sqrt{\frac{1}{N} \sum_{i=1}^{N} \|f_i\|_{L^2}^4} \leq L^2,$$

using that $\|f\|_{L^2} \leq \|f\|_{C^s} \leq L$.

It remains to compute the integral above. By Proposition 7 below, we have an upper bound on the covering number of $\mathcal{H}_J$, which translates to a bound on the packing numbers by the inequality $D(\mathcal{H}_J, \rho, \tau) \leq N(\mathcal{H}_J, \rho, \tau/2)$. Using that and equation (24) yields

$$\int_0^{\sqrt{\|P_N h^2\|_{\mathcal{H}_J}}} \sqrt{\log 2 N(\mathcal{H}_J, \tilde{\rho}_N, \tau/2)} \, d\tau$$
$$\leq \int_0^B \sqrt{\log 2 + H_J + 2^{3(d+1)} (\operatorname{diam}(\mathbb{M}))^d C_{\psi, H^r} \left(C_W 4\sqrt{B}/\delta\right)^{d/r} + 2^{d+1} J S_{\text{filter}} \log(C_W 4\sqrt{B}/\delta)} \, d\tau.$$

where the constants $H_J$ and $C_W$ are defined in the statement of Proposition 7. Applying Claim 1 to bound the integral yields

$$\int_0^{\sqrt{\|P_N h^2\|_{\mathcal{H}_J}}} \sqrt{\log 2 N(\mathcal{H}_J, \tilde{\rho}_N, \tau/2)} \, d\tau$$
$$\leq B \bigg( \sqrt{\log 2 + H_J} + \sqrt{2^{3(d+1)} \operatorname{diam}(\mathbb{M})^{d^2/r} C_{\psi,H^r}^{d/r}} \frac{(C_W 4\sqrt{B}/B)^{d/(2r)}}{1 - d/(2r)}$$
$$+ 2\sqrt{2^{d+1} J \, S_{\text{filter}} \log(C_W 4\sqrt{B}/B)} \bigg).$$

Using that $B \geq 2^{Jd/2+d} 4 C_{\psi,L^2}^2 J^2 \max\{LC_{\mathcal{T}}, \sigma\}^2$, we see that

$$\frac{4C_W}{\sqrt{B}} \leq \frac{4 \cdot 26 \cdot J^2 \cdot 2^{Jd/2+d} C_{\psi,L^2} \max\{LC_{\mathcal{T}}, \sigma, 1\}}{2 C_{\psi,L^2} \cdot J \max\{LC_{\mathcal{T}}, \sigma\} \, 2^{d+dJ/2}} \leq 52 \, J.$$

The expression above can hence be bounded by

$$\int_0^{\sqrt{\|P_N h^2\|_{\mathcal{H}_J}}} \sqrt{\log 2 N(\mathcal{H}_J, \tilde{\rho}_N, \tau/2)} \, d\tau$$
$$\leq B \bigg( \sqrt{\log 2 + H_J} + \sqrt{2^{3(d+1)} \operatorname{diam}(\mathbb{M})^{d^2/r} C_{\psi,H^r}^{d/r}} \frac{(52\, J)^{d/(2r)}}{1 - d/(2r)}$$
$$+ 2\sqrt{2^{d+1} J \, S_{\text{filter}} \log(52\, J)} \bigg).$$

Defining $C_B = 18 \cdot 2^{2d} (L^2 C_{\mathcal{T}}^2 + \sigma^2)$, we have $B \leq C_B C_\psi^2 J^2 \, 2^{Jd}$. This yields

$$\int_0^{\sqrt{\|P_N h^2\|_{\mathcal{H}_J}}} \sqrt{\log 2 N(\mathcal{H}_J, \tilde{\rho}_N, \tau/2)} \, d\tau$$
$$\leq C_B \, 2^{Jd} \, C_{\psi,L^2}^2 \, J^3 \bigg[ \sqrt{\frac{\log 2 + H_J}{J^2}} + 2^{3(d+1)/2} \frac{52^{d/(2r)}}{1 - d/(2r)} (\operatorname{diam}(\mathbb{M})^d \, C_{\psi,H^r})^{d/(2r)}$$
$$+ 2\sqrt{2^{d+1} S_{\text{filter}} \frac{\log(52\, J)}{J}} \bigg].$$

For $J$ large enough, there is a constant $C_0$ such that

$$\sqrt{\frac{\log 2 + H_J}{J^2}} + 2\sqrt{2^{d+1} \frac{\log(52\, J)}{J}} \leq C_0.$$

Define further the constant $C_1 = 2^{3(d+1)/2} \frac{52^{d/(2r)}}{1 - d/(2r)}$. This implies that

$$\int_0^{\sqrt{\|P_N h^2\|_{\mathcal{H}_J}}} \sqrt{\log 2 N(\mathcal{H}_J, \tilde{\rho}_N, \tau/2)} \, d\tau$$

$$\leq C_B\, 2^{Jd}\, C_{\psi,L^2}^2\, J^3 \left( S_{\text{filter}}\, C_0 + C_1 (\text{diam}(\mathbb{M})^d\, C_{\psi,H^r})^{d/(2r)} \right).$$

Putting everything together yields

$$\mathbb{E}_{f_i} \sup_{F \in \mathcal{F}_J} \left| \frac{1}{N} \sum_{i=1}^N h_F(f_i) - \mathbb{E}_f h_F(f) \right|$$

$$\leq \frac{2L^2}{\sqrt{N}} + \frac{8\sqrt{2}}{\sqrt{N}}\, C_B\, 2^{Jd}\, C_{\psi,L^2}^2\, J^3 \left( S_{\text{filter}}\, C_0 + C_1 (\text{diam}(\mathbb{M})^d\, C_{\psi,H^r})^{d/(2r)} \right)$$

$$\leq \overline{C} \frac{C_{\psi,L^2}^2\, J^3\, 2^{Jd}}{\sqrt{N}}\, \max\{S_{\text{filter}}, (\text{diam}(\mathbb{M})^d\, C_{\psi,H^r})^{d/(2r)}\}$$

where

$$\overline{C} = 2L^2 + 8\sqrt{2}(C_1 + C_0),$$

which yields the claim. □

**Claim 1.** For $K \geq 1$, $A, a, \alpha, C > 0$, $r > d/2$ and $C \geq e\, A$, we have

$$\int_0^A \sqrt{K + a\, (C/\tau)^{d/r} + \alpha \log C/\tau}\, d\tau \leq A \left( \sqrt{K} + \sqrt{a}\, \frac{(C/A)^{d/(2r)}}{1 - d/(2r)} + 2\sqrt{\alpha \log C/A} \right).$$

*Proof.* For $K, x \geq 0$ we have $\sqrt{K + x} \leq \sqrt{K} + \sqrt{x}$. This implies that

$$\int_0^A \sqrt{K + a\, (C/\tau)^{d/r} + \alpha \log C/\tau}\, d\tau = \frac{C}{\alpha} \int_{\alpha \log A/C}^\infty \sqrt{K + a\, e^{xd/(\alpha r)} + x}\, e^{-x/\alpha}\, dx$$

$$= \frac{C}{\alpha} \int_{\alpha \log C/A}^\infty \left( \sqrt{K} e^{-x/\alpha} + \sqrt{a} e^{xd/(2\alpha r) - x/\alpha} + \sqrt{x} e^{-x/\alpha} \right) dx$$

$$= \sqrt{K} A + \sqrt{a} C \frac{e^{-(1 - d/(2r)) \log C/A}}{1 - d/(2r)} + C\sqrt{\alpha} \int_{\log C/A}^\infty \sqrt{x} e^{-x}\, dx.$$

Now we have

$$\int_{\log C/A}^\infty \sqrt{x} e^{-x}\, dx \leq \frac{1}{\sqrt{\log C/A}} \int_{\log C/A}^\infty x e^{-x}\, dx = \frac{1 + \log C/A}{\sqrt{\log C/A}} e^{-\log C/A}$$

$$\leq 2\sqrt{\log C/A}\, \frac{A}{C}$$

using that $\int_a^\infty x\, e^{-x}\, dx = (a + 1)e^{-a}$. This completes the proof. □

### 4.3.1. Covering numbers of $\mathcal{H}_J$

Given a set of independent random functions $f_i \sim \Pi$, define a random semi-metric in $\mathcal{F}_J$ by

$$\rho_N(F_1, F_2) := \sqrt{N^{-1} \sum_{i=1}^N \mathbb{E}_{W_i} \|F_1(Tf_i + \sigma\, dW_i) - F_2(Tf_i + \sigma\, dW_i)\|^2}$$

In this subsection, we bound the covering numbers of $\mathcal{H}_J$ with respect to $\tilde{\rho}_N$ by upper bounding them with the covering numbers of $\mathcal{F}_J$ with respect to the semi-metric $\rho_N$.

**Proposition 7.** *The covering numbers of $\mathcal{H}_J(r, R, S_{\text{filter}}, \kappa_\tau, C_{\psi,L^2}, C_{\psi,H^r})$ satisfy*

$$\log N\big(\mathcal{H}_J, \tilde{\rho}_N, \delta\big) \leq H_J + 2^{3(d+1)} \big(\operatorname{diam}(\mathbb{M})^d C_{\psi,H^r} C_W 2\sqrt{B}/\delta\big)^{d/r}$$
$$+ 2^{d+1} J S_{\text{filter}} \log(C_W 2\sqrt{B}/\delta).$$

*where* $H_J = 2^d J S_{\text{filter}} \log(3 \max\{1, \kappa_\tau 2^{Jd/2}\})$, $C_W = C_{\psi,L^2} 26 J^2 2^{Jd/2+d} \max\{C_\mathcal{T} L, \sigma, 1\}$ *and* $B = 2L^2 + 4C_{\psi,L^2}^2 (J+1)^2 (L^2 C_\mathcal{T}^2 + \sigma^2)$, *where* $\|f_i\| \leq L$ *and* $\|T\|_{op} \leq C_\mathcal{T}$.

*Proof.* By Claim 3 we have the inequality

$$\log N(\mathcal{H}_J, \tilde{\rho}_N, \delta) \leq \log N(\mathcal{F}_J, \rho_N, \delta/(2\sqrt{B})),$$

and the right-hand side can be bounded with Proposition 9 to yield the result. □

**Claim 2.** *For any $h \in \mathcal{H}_J$ given by $h(f) = \mathbb{E}_W \|F(Tf + \sigma\, dW) - f\|_{L^2}^2$ for $F$ with filters $\phi, \psi$ and $\alpha^{(j)}, a^{(j)}$, etc., as in Definition 5 we have the bound*

$$|h(f)| \leq 2\|f\|_{L^2}^2 + 4 \cdot 2^{Jd} \|\psi\|_{L^2}^2 (2^d J + 1)^2 \big(\|f\|_{L^2}^2 \|T\|_{op}^2 + \sigma^2\big).$$

*Proof.* Using the definition of $h$ and the bounds in Proposition 10 for bounding $\|\bar{s}^J\|$, we get

$$|h(f)| \leq \mathbb{E}_W \big(\|f\| + \|\phi\| \|\bar{s}^{(J)}\|\big)^2$$
$$\leq 2\|f\|^2 + 2 \cdot 2^{Jd} \|\phi\|^2 (2^d J + 1)^2 \mathbb{E}_W \|\langle Tf + \sigma\, dW, \psi\rangle\|^2$$
$$\leq 2\|f\|^2 + 4 \cdot 2^{Jd} \|\phi\|^2 (2^d J + 1)^2 \mathbb{E}_W \big(\|\langle Tf, \psi\rangle\|^2 + \|\langle \sigma\, dW, \psi\rangle\|^2\big)$$
$$\leq 2\|f\|^2 + 4 \cdot 2^{Jd} \|\phi\|^2 (2^d J + 1)^2 \big(\|Tf\|^2 \|\psi\|^2 + \sigma^2 \|\psi\|^2\big)$$
$$\leq 2\|f\|^2 + 4 \cdot 2^{Jd} \|\phi\|^2 (2^d J + 1)^2 \big(\|f\|^2 \|T\|^2 + \sigma^2\big) \|\psi\|^2 \square$$

**Claim 3.** *Consider the random semi-metrics $\tilde{\rho}_N$ in $\mathcal{H}_J(r, R, S_{\text{filter}}, \kappa_\tau, C_{\psi,L^2}, C_{\psi,H^r})$ and $\rho_N$ in $\mathcal{F}_J(r, R, S_{\text{filter}}, \kappa_\tau, C_{\psi,L^2}, C_{\psi,H^r})$ based on the same set of random functions $f_i \sim \Pi$. Then, conditionally on the functions $f_i$, we have*

$$\log N(\mathcal{H}_J, \tilde{\rho}_N, 2\sqrt{B}\delta) \leq \log N(\mathcal{F}_J, \rho_N, \delta)$$

*for* $B = 2L^2 + 4 \cdot 2^{Jd} C_{\psi,L^2}^2 (2^d J + 1)^2 (L^2 C_\mathcal{T}^2 + \sigma^2)$, *for* $\|f\| \leq L$ *and* $\|T\|_{op} \leq C_\mathcal{T}$.

*Proof.* Note that

$$|h_{F_1}(f_i) - h_{F_2}(f_i)|$$
$$= |\mathbb{E}\big(\|F_1(Tf_i + \sigma\, dW) - f_i\|^2 - \|F_2(Tf_i + \sigma\, dW) - f_i\|^2\big)|$$

$$\begin{aligned}
&= |\mathbb{E}\big(\|F_1(Tf_i + \sigma\,dW) - f_i\| - \|F_2(Tf_i + \sigma\,dW) - f_i\|\big) \\
&\quad \times \big(\|F_1(Tf_i + \sigma\,dW) - f_i\| + \|F_2(Tf_i + \sigma\,dW) - f_i\|\big)| \\
&\leq \sqrt{\mathbb{E}\big(\|F_1(Tf_i + \sigma\,dW) - f_i\| - \|F_2(Tf_i + \sigma\,dW) - f_i\|\big)^2} \\
&\quad \times \sqrt{\mathbb{E}\big(\|F_1(Tf_i + \sigma\,dW) - f_i\| + \|F_2(Tf_i + \sigma\,dW) - f_i\|\big)^2} \\
&\leq \sqrt{\mathbb{E}\|F_1(Tf_i + \sigma\,dW) - F_2(Tf_i + \sigma\,dW)\|^2}\, \underbrace{\sqrt{2(h_{F_1}(f_i) + h_{F_2}(f_i))}}_{\leq 2\sqrt{B}},
\end{aligned}$$

where in the last inequality we use Claim 2. Hence, we have

$$\begin{aligned}
\tilde{\rho}_N(h_{F_1}, h_{F_2}) &\leq \sqrt{N^{-1} \sum_{i=1}^N 4B\mathbb{E}\|F_1(Tf_i + \sigma\,dW) - F_2(Tf_i + \sigma\,dW)\|^2} \\
&\leq 2\sqrt{B}\,\rho_N(F_1, F_2),
\end{aligned}$$

which yields the claim. □

### 4.3.2. Covering numbers of $\mathcal{F}_J$

**Proposition 8.** *Let $F, G$ be networks in $\mathcal{F}_J$, and denote the differences between their filters $a^{(k)}$ by $\Delta^{(a,k)}$, between their filters $\psi$ by $\Delta^{(\psi)}$, and so on. Then*

$$\|F(f) - G(f)\|_{L^2} \leq |\langle f, \Delta^{(\psi)}\rangle_{L^2}|\, 2^{Jd/2}(1 + 2^d J) + 2^{Jd/2} 2^d \sum_{k=0}^{J-1} |\Delta^{(\tau,k)}| \quad (25)$$
$$+ |\langle f, \psi\rangle_{L^2}|\, 2^{Jd/2}\bigg(\sum_{k=1}^J (1 + 2^d k)\|\Delta^{(\alpha,k)}\|_{\ell^2} + (1 + 2^d k)\|\Delta^{(a,k)}\|_{\ell^2}$$
$$+ \sum_e \|\Delta^{(b,k,e)}\|_{\ell^2} + \|\Delta^{(\beta,k,e)}\|_{\ell^2}\bigg).$$

*Proof.* We construct a sequence $\{F_l, l = 0, \ldots, E\}$ of SU-nets contained in $\mathcal{F}_J$ that begins at $F_0 = F$, ends at $F_E = G$, and each pair of consecutive nets only differ in one filter. Since there are $1 + 2^{d+1}J$ filters and $J$ thresholds in an SU-net, we can achieve this with a sequence of $E = 1 + (2^{d+1} + 1)J + 1$ nets (the "+1" being $F_0 = F$).

With this construction, we have

$$\|F(f) - G(f)\|_{L^2} \leq \sum_{l=0}^{E-1} \|F_{l+1}(f) - F_l(f)\|_{L^2}$$
$$=: P_\psi + \sum_{k=1}^J P_{\alpha,k} + P_{a,k} + \sum_e P_{b,k,e} + P_{\beta,k,e} + P_{\tau,k-1},$$

where $P_\psi$ denotes the perturbation in the filter $\psi$, $P_{\alpha,k}$ in filter $\alpha^{(k)}$, and accordingly for the other terms. By Proposition 11 below, each of those terms can be bounded to yield the claim. □

**Proposition 9.** Let $J, S_{\text{filter}} \in \mathbb{N}$ and $r, \kappa_\tau, C_{\psi,L^2}, C_{\psi,H^r} > 0$ be fixed. The covering numbers of $\mathcal{F}_J(r, R, S_{\text{filter}}, \kappa_\tau, C_{\psi,L^2}, C_{\psi,H^r})$ with respect to the semimetric $\tilde{\rho}_N$ satisfy

$$\log N(\mathcal{F}_J, \tilde{\rho}_N, \delta)$$
$$\leq H_J + 2^{3(d+1)}\big(\operatorname{diam}(\mathbb{M})^d C_{\psi,H^r} C_W/\delta\big)^{d/r} + 2^{d+1} J S_{\text{filter}} \log(C_W/\delta).$$

where $H_J = 2^d J S_{\text{filter}} \log(3 \max\{1, \kappa_\tau 2^{Jd/2}\})$ and
$C_W = C_{\psi,L^2} 26 J^2 2^{Jd/2+d} \max\{C_\mathcal{T} L, \sigma, 1\}$.

*Proof.* First notice that we can cover a ball of Euclidean radius $R$ in $\mathbb{R}^s$ with $(3R/\epsilon)^s$ balls of radius $\epsilon$ (see e.g. Proposition 4.3.34 in [19]). Using that the discrete filters in Definition 5 have support $S_{\text{filter}}$ and Euclidean norm bounded by one, this means that we can cover the set of all possible filters in $\mathcal{F}_J$ with

$$N_{filters} := (3/\epsilon)^{2^d J S_{\text{filter}}} \cdot e^{2^{3(d+1)}\big(\operatorname{diam}(\mathbb{M})^d C_{\psi,H^r}/\epsilon\big)^{d/r}}$$

filters whose Euclidean distance to each other is not more that $\epsilon$. The second term arises from the continuous filter $\psi$, which belongs to the ball of radius $C_{\psi,H^r}$ in the Sobolev space $H^r(\mathbb{M})$ (see Definition 5). Theorem 4.3.36 in [19] then gives the bound on the covering numbers of the unit ball in $H^r$. The constant $2^{3(d+1)}$ is justified in Lemma 1 below, while the diameter of $\mathbb{M}$ appears by rescaling. On the other hand, we cover the set of possible thresholds $\tau_j$ in a different way: each threshold $\tau_j$ takes values in $[0, \kappa_\tau]$, and we cover this range with balls of radius $\epsilon 2^{-Jd/2}$. This yields a number

$$N_{thres} := \max\{1, (\kappa_\tau 2^{Jd/2}/\epsilon)^J\}$$

of centers needed to cover all the thresholds.

For $i = 1, \ldots, N$, define now the semimetric $d_i(F, G)$ on $\mathcal{F}_J$ by

$$d_i(F, G) = \mathbb{E}_{W_i} \|F(Tf_i + \sigma dW_i) - G(Tf_i + \sigma dW_i)\|_{L^2}.$$

Using equation (25) in Proposition 8, this implies that we can construct $N_{filters} \cdot N_{thres}$ SU-nets that cover $\mathcal{F}_J$ with $d_i$-balls of radius

$$\mathbb{E}_{W_i}\left[|\langle Tf_i + \sigma dW_i, \psi\rangle_{L^2}| \, 2 \, 2^{Jd/2} \sum_{k=1}^{J} 2^d (1+k) + 2^{Jd/2} 2^d J 2^{-Jd/2}\right] \epsilon$$
$$+ \mathbb{E}_{W_i} |\langle Tf_i + \sigma dW_i, \Delta^{(\psi)}\rangle_{L^2}| 2^{Jd/2}(1 + 2^d J)$$
$$\leq \left[(\|Tf_i\|_{L^2} C_{\psi,L^2} + \sigma C_{\psi,L^2}\sqrt{2/\pi}) \, 2 \sum_{k=1}^{J}(1+k) + (\|Tf_i\|_{L^2} + \sigma\sqrt{2/\pi}) \, 2J + J\right]$$
$$\times 2^{Jd/2+d} \epsilon$$

$$\leq C_{\psi,L^2}\left[8J^2+4\,J+J\right]\max\{C_\mathcal{T}L,\sigma,1\}\,2^{Jd/2+d+1}\epsilon,$$

using that $\mathbb{E}|X|=\sqrt{2/\pi}$ for $X\sim\mathcal{N}(0,1)$, and that $\|f_i\|_{L^2}\leq\|f_i\|_{C^s}\leq L$ and that $\|T\|_{op}\leq C_\mathcal{T}$. This means that we have $N_{filters}\cdot N_{thres}$ SU-nets that cover $\mathcal{F}_J$ with $\tilde\rho_N$-balls of radius

$$C_{\psi,L^2}\,26\,J^2\,2^{Jd/2+d}\,\max\{C_\mathcal{T}L,\sigma,1\}\epsilon$$

Defining $\delta:=C_{\psi,L^2}\,26\,J^2\,2^{Jd/2+d}\,\max\{C_\mathcal{T}L,\sigma,1\}\epsilon$, we can reformulate the above statement as follows: we can cover $\mathcal{F}_J$ with

$$N_\delta:=3^{2^dJ\,S_\text{filter}}\cdot\exp\{2^{3(d+1)}(C_{\psi,H^r}C_{\psi,L^2}\,26\,J^2\,2^{Jd/2+d}\,\max\{C_\mathcal{T}L,\sigma,1\}/\delta)^{d/r}\}$$
$$\cdot(C_{\psi,L^2}\,26\,J^2\,2^{Jd/2+d}\,\max\{C_\mathcal{T}L,\sigma,1\}/\delta)^{2^dJ\,S_\text{filter}}\cdot\max\{1,(\kappa_\tau\,2^{Jd/2})^J\}$$

balls of radius $\delta$ with respect to the semimetric $\tilde\rho_N$. The logarithm of this number can be further bounded as

$$\log N_\delta\leq 2^dJ\,S_\text{filter}\log(3\max\{1,\kappa_\tau\,2^{Jd/2}\})$$
$$+2^{3(d+1)}\bigl(C_{\psi,H^r}C_{\psi,L^2}\,26\,J^2\,2^{Jd/2+d}\,\max\{C_\mathcal{T}L,\sigma,1\}/\delta\bigr)^{d/r}$$
$$+2^{d+1}J\,S_\text{filter}\log(C_{\psi,L^2}\,26\,J^2\,2^{Jd/2+d}\,\max\{C_\mathcal{T}L,\sigma,1\}/\delta)$$

This yields the claim. □

**Lemma 1.** *Let $r\in\mathbb{N}$ satisfy $r>d/2$. Denote the unit ball of a Sobolev space $H^t$ by $H_1^t$. The covering numbers of $H_1^r$ in $L^2$ satisfy*

$$\log N(H_1^r,L^2,\epsilon)\leq 2^{3(d+1)}\,\epsilon^{-d/r}$$

*for any $\epsilon\in(0,1)$.*

*Proof.* The result follows from Theorem 4.3.36 in [19], where the constant is not specified. Here we give an upper bound for the constant. First notice that the conditions on $r$ imply that $r\geq 1$. Hence, the unit ball of $H^r$ is contained in the unit ball of $H^1$, so we have

$$\log N(H_1^r,L^2,1)\leq\log N(H_1^1,L^2,1)\leq K_{H^1}.$$

We hence have reduced the problem to computing the constant $K_{H^1}$ instead of $K_{H^r}$. We proceed as in the proof of Theorem 4.3.36 in [19], but keeping track of the constants.

In order to bound the covering numbers of $H_1^1$ with balls of radius $\epsilon$, let $j_0=\lfloor\log\epsilon^{-1}\rfloor$. For any two functions $f^{(a)},f^{(b)}\in H_1^1$, we have

$$\|f^{(a)}-f^{(b)}\|_{L^2}^2=\sum_{j\leq j_0}\sum_{k,e}|c_{j,k,e}^{(a)}-c_{j,k,e}^{(b)}|^2+\sum_{j>j_0}\sum_{k,e}|c_{j,k,e}^{(a)}-c_{j,k,e}^{(b)}|^2$$
$$\leq\sum_{j\leq j_0}\sum_{k,e}|c_{j,k,e}^{(a)}-c_{j,k,e}^{(b)}|^2+\sum_{j>j_0}\sum_{k,e}|c_{j,k,e}^{(a)}|^2+|c_{j,k,e}^{(b)}|^2$$

$$\leq \sum_{j\leq j_0}\sum_{k,e}|c_{j,k,e}^{(a)} - c_{j,k,e}^{(b)}|^2 + 2^{-2j_0}\sum_{j>j_0}\sum_{k,e}2^{2j}(|c_{j,k,e}^{(a)}|^2 + |c_{j,k,e}^{(b)}|^2)$$

$$\leq \sum_{j\leq j_0}\sum_{k,e}|c_{j,k,e}^{(a)} - c_{j,k,e}^{(b)}|^2 + 2^{-2j_0+1}$$

using that $f^{(a)}, f^{(b)} \in H^1$, whence

$$\sum_{j,k,e}2^{2j}|c_{j,k,e}^{(a)}|^2 \leq \|f^{(a)}\|_{H^1} \leq 1.$$

For every $j \in \{0, \ldots, j_0\}$, choose an $\epsilon'_j$ covering of $b_2^{2^{jd}(2^d-1)}$, the ball of $\ell^2$-radius one in dimension $2^{jd}(2^d - 1)$ (i.e. on the $k$ and $e$ coefficients). By Proposition 4.3.34 in [19], this can be done with at most $(3\,\epsilon^{-1})^{2^{dj}(2^d-1)}$. Notice that, by the equations above, the $\epsilon'_j$ coverings induce a covering of $H_1^1$ with balls of radius $R = \sum_{j\leq j_0} 2^{-j}\epsilon'_j + 2^{-j_0+1}$. Combining this with the above yields

$$\log N(H_1^1, L^2, R) \leq \sum_{j=0}^{j_0}\log N(b_2^{2^{jd}(2^d-1)}, \ell^2, \epsilon'_j) \leq \sum_{j=0}^{j_0}2^{dj}(2^d - 1)\log(3\epsilon'_j{}^{-1}).$$

Choosing now $\epsilon'_j = 2^{2(j-j_0)}$ yields $R = 4\, 2^{-j_0} \leq 4\,\epsilon$ and

$$\log N(H_1^1, L^2, R) \leq 2^d \sum_{j=0}^{j_0} 2^{dj}\big(\log 3 + 2(j_0 - j)\log 2\big)$$

$$\leq 2^d \log(3)\, 2^{dj_0}\, 2 + 2^{d+1}\log(2)\, 2^{dj_0}\underbrace{\sum_{j=0}^{j_0}2^{d(j-j_0)}(j_0 - j)}_{\leq 2}$$

$$\leq 2^{dj_0}\, 2^{d+1}\log 12.$$

Plugging in the values for $R$ and $j_0$ gives

$$\log N(H_1^1, L^2, \epsilon) \leq \epsilon^{-d}\, 4^d\, 2^{d+1}\log 12 \leq \epsilon^{-d}\, 2^{3(d+1)},$$

which gives the claim. $\square$

### 4.4. Stability of F under perturbations

In this section we prove stability estimates for networks under perturbations of their filters.

For simplicity of the notation, all norms $\|\cdot\|$ in this sections are the $\ell^2$ norm unless otherwise stated.

**Proposition 10** (Size of the terms). Using the notation and assumptions from Definition 5 for a network $F$ with input $f$, the following estimates for the inner layers hold:

$$\|s^{(J-l)}\| \leq 2^{Jd/2}\,|\langle f, \psi\rangle_{L^2}|$$

$$\|d^{(J-l)}\| \leq 2^{Jd/2} |\langle f, \psi \rangle_{L^2}|$$
$$\|\overline{d}^{(J-l)}\| \leq \|d^{(J-l)}\|$$
$$\|\overline{s}^{(J-l)}\| \leq \left(1 + 2^d(J-l)\right) 2^{Jd/2} |\langle f, \psi \rangle_{L^2}|$$

for any $l = 0, \ldots, J$.

*Proof.* For the first inequality, using the definition of $s^{(J-l)}$ in Definition 4 we have

$$\|s^{(J-l)}\| = \left(\sum_n \left|\sum_m \alpha_m^{(J-l)} s_{2n-m}^{(J-l+1)}\right|^2\right)^{1/2} \leq \left(\sum_n \sum_m |\alpha_m^{(J-l)}|^2 \sum_m |s_{2n-m}^{(J-l+1)}|^2\right)^{1/2}$$
$$\leq \|\alpha^{(J-l)}\| \left(\sum_n \sum_m |s_{2n-m}^{(J-l+1)}|^2\right)^{1/2} = \|\alpha^{(J-l)}\| \|s^{(J-l+1)}\| \leq \|s^{(J-l+1)}\|,$$

where we use that $\|\alpha^{(l)}\| \leq 1$ by Definition 5. Iterating this, we conclude that

$$\|s^{(J-l)}\| \leq \|s^{(J)}\| = \sqrt{\sum_k |\langle \psi_k, f \rangle|} \leq 2^{Jd/2} |\langle f, \psi \rangle_{L^2}|$$

for any $l = 0, \ldots, J$.

For the second term, the proof proceeds analogously, with the obvious change that yields a factor $\beta^{(J-l),e}$ instead of $\alpha^{(J-l)}$ in the last iteration.

The proof of the third inequality is trivial due to the Lipschitz property of $\rho$ and of the nonnegativity of $\tau_j$.

For the last inequality, we use the definition in (8) of the coefficients

$$\|\overline{s}^{(J-l)}\| \leq \|a^{(J-l-1)}\| \|\overline{s}^{(J-l-1)}\| + \sum_e \|b^{(J-l-1),e}\| \|\overline{d}^{(J-l-1),e}\|$$
$$\leq \|\overline{s}^{(J-l-1)}\| + \sum_e \|\overline{d}^{(J-l-1),e}\|$$

Iterating this yields

$$\|\overline{s}^{(J-l)}\| \leq \|\overline{s}^{(0)}\| + \sum_{h=0}^{J-l-1} \sum_e \|\overline{d}^{(h),e}\|.$$

Applying now the first inequalities proven above, we get

$$\|\overline{s}^{(J-l)}\| \leq \left(1 + 2^d(J-l)\right) 2^{Jd/2} |\langle f, \psi \rangle_{L^2}| \square$$

**Proposition 11** (Perturbation of filters). Let $F$ be an SU-net and $F_\Delta$ be the same net with one of the filters modified by adding a perturbation $\Delta$. Then

$$\|F(f) - F_\Delta(f)\|_{L^2} \leq c_k \|\Delta\| \|\phi\|_{L^2} 2^{Jd/2} |\langle f, \psi \rangle_{L^2}|,$$

where the constant $c_k$ equals 1 if the perturbation is on the coefficients $b^{(k),e}$ or $\beta^{(k),e}$, and $1 + 2^d k$ if the perturbation is on the coefficients $a^{(k)}$ or $\alpha^{(k)}$.

If the filters $\phi$ or $\psi$ are perturbed, the result is

$$\|F(f) - F_\Delta(f)\|_{L^2} \leq (1 + 2^d J)\|\Delta\|_{L^2} 2^{Jd/2} \|\psi\|_{L^2} \|f\|_{L^2} \quad \text{for perturbed } \phi$$
$$\|F(f) - F_\Delta(f)\|_{L^2} \leq (1 + 2^d J)\|\phi\|_{L^2} 2^{Jd/2} |\langle f, \Delta\rangle_{L^2}| \quad \text{for perturbed } \psi.$$

If the threshold $\tau_k$ is perturbed by a number $\Delta$, the result is

$$\|F(f) - F_\Delta(f)\|_{L^2} \leq 2^{Jd/2} 2^d \|\phi\|_{L^2} |\Delta|.$$

*Proof.* **Perturbation of** $a^{(k)}$. We begin with the perturbation of the $a^{(k)}$ filters. By the definition notice that the perturbation $a_m^{(k)} \mapsto a_m^{(k)} + \Delta_m$ induces a change in $F(f)$ to $F(f) + \xi$ where

$$\xi = \sum_n \phi_n(x) \sum_{r_0} a_{r_0}^{(J-1)} \sum_{r_1} a_{r_1}^{(J-2)} \cdots \sum_{r_{J-k-1}} \Delta_{r_{J-k-1}} \overline{s}_{2^{-J+k}n + 2^{-J+k}r_0 + \cdots + 2^{-1}r_{J-k-1}}^{(k)}.$$

Since $\phi_k(x)$ is an orthonormal wavelet basis by Definition 5, we have that

$$\int |\sum_k \phi_k(x) A_k|^2\, dx = \sum_{k_1, k_2} A_{k_1} A_{k_2} \underbrace{\int \phi_{k_1}(x) \phi_{k_2}(x)\, dx}_{=\delta_{k_1,k_2} \|\phi\|_{L^2}^2} = \|\phi\|_{L^2}^2 \sum_k |A_k|^2.$$

Hence we have

$$\|F(f) - F_\Delta(f)\|$$
$$= \left( \int \left| \sum_n \phi_n(x) \sum_{r_0} a_{r_0}^{(J-1)} \sum_{r_1} a_{r_1}^{(J-2)} \cdots \right.\right.$$
$$\left.\left. \times \sum_{r_{J-k-1}} \Delta_{r_{J-k-1}} \overline{s}_{2^{-J+k}n + 2^{-J+k}r_0 + \cdots + 2^{-1}r_{J-k-1}}^{(k)} \right|^2 dx \right)^{1/2}$$
$$\leq \left( \|\phi\|_{L^2}^2 \sum_n \left| \sum_{r_0} a_{r_0}^{(J-1)} \sum_{r_1} a_{r_1}^{(J-2)} \cdots \right.\right.$$
$$\left.\left. \times \sum_{r_{J-k-1}} \Delta_{r_{J-k-1}} \overline{s}_{2^{-J+k}n + 2^{-J+k}r_0 + \cdots + 2^{-1}r_{J-k-1}}^{(k)} \right|^2 \right)^{1/2}$$
$$\leq \|\phi\|_{L^2} \left( \sum_n \underbrace{\sum_{r_0} |a_{r_0}^{(J-1)}|^2}_{\|a^{(J-1)}\|^2} \sum_{r_0} \left| \sum_{r_1} a_{r_1}^{(J-2)} \cdots \right.\right.$$
$$\left.\left. \times \sum_{r_{J-k-1}} \Delta_{r_{J-k-1}} \overline{s}_{2^{-J+k}n + 2^{-J+k}r_0 + \cdots + 2^{-1}r_{J-k-1}}^{(k)} \right|^2 \right)^{1/2}$$
$$\vdots$$
$$\leq \|\phi\|_{L^2} \prod_{l=1}^{J-k-1} \|a^{(J-l)}\| \left( \sum_n \sum_{r_0} \cdots \sum_{r_{J-k-2}} \right.$$

$$\left| \sum_{r_{J-k-1}} \Delta_{r_{J-k-1}} \overline{s}^{(k)}_{2^{-J+k}n + 2^{-J+k}r_0 \cdots + 2^{-1}r_{J-k-1}} \right|^2 \Bigg)^{1/2}$$

$$\leq \|\phi\|_{L^2} \underbrace{\prod_{l=1}^{J-k-1} \|a^{(J-l)}\|}_{\leq 1} \Bigg( \sum_{r_{J-k-1}} |\Delta_{r_{J-k-1}}|^2 \sum_n \sum_{r_0} \cdots$$

$$\sum_{r_{J-k-2}} \sum_{r_{J-k-1}} \left| \overline{s}^{(k)}_{2^{-J+k}n + 2^{-J+k}r_0 \cdots + 2^{-1}r_{J-k-1}} \right|^2 \Bigg)^{1/2}$$

$$\leq \|\phi\|_{L^2} \|\Delta\| \|\overline{s}^{(k)}\|$$

$$\leq (1 + 2^d \, k) \, \|\phi\|_{L^2} \|\Delta\| \, 2^{Jd/2} \, |\langle f, \psi \rangle_{L^2}|$$

using Proposition 10 in the last inequality.

**Perturbation of** $b^{(k),e}$. This case is analogous, since we use the fact that the perturbation $b^{(k),e}_m \mapsto b^{(k),e}_m + \Delta_m$ induces the change $F(f) \mapsto F(f) + \xi$ for

$$\xi = \sum_n \phi_n(x) \sum_{r_0} a^{(J-1)}_{r_0} \sum_{r_1} a^{(J-2)}_{r_1} \cdots$$

$$\times \sum_{r_{J-k-2}} a^{(k+1)}_{r_{J-k-2}} \sum_{r_{J-k-1}} \Delta_{r_{J-k-1}} \overline{d}^{(k),e}_{2^{-J+k}n + 2^{-J+k}r_0 \cdots + 2^{-1}r_{J-k-1}}.$$

Defining

$$A^{(k),e}_{r_0,\ldots,r_{J-k-1}} := \overline{d}^{(k),e}_{2^{-J+k}n + 2^{-J+k}r_0 \cdots + 2^{-1}r_{J-k-1}}$$

In the same way as above, this yields

$$\|F(f) - F_\Delta(f)\|$$

$$= \Bigg( \sum_x \bigg| \sum_n \phi_n(x) \sum_{r_0} a^{(J-1)}_{r_0} \sum_{r_1} a^{(J-2)}_{r_1} \cdots$$

$$\times \sum_{r_{J-k-2}} a^{(k+1)}_{r_{J-k-2}} \sum_{r_{J-k-1}} \Delta_{r_{J-k-1}} A^{(k),e}_{r_0,\ldots,r_{J-k-1}} \bigg|^2 \Bigg)^{1/2}$$

$$\leq \Bigg( \|\phi\|^2_{L^2} \sum_n \bigg| \sum_{r_0} a^{(J-1)}_{r_0} \sum_{r_1} a^{(J-2)}_{r_1} \cdots$$

$$\times \sum_{r_{J-k-2}} a^{(k+1)}_{r_{J-k-2}} \sum_{r_{J-k-1}} \Delta_{r_{J-k-1}} A^{(k),e}_{r_0,\ldots,r_{J-k-1}} \bigg|^2 \Bigg)^{1/2}$$

$$\leq \|\phi\|_{L^2} \Bigg( \sum_n \underbrace{\sum_{r_0} |a^{(J-1)}_{r_0}|^2}_{\|a^{(J-1)}\|^2} \sum_{r_0} \bigg| \sum_{r_1} a^{(J-2)}_{r_1} \cdots$$

$$\times \sum_{r_{J-k-2}} a^{(k+1)}_{r_{J-k-2}} \sum_{r_{J-k-1}} \Delta_{r_{J-k-1}} A^{(k),e}_{r_0,\ldots,r_{J-k-1}} \bigg|^2 \Bigg)^{1/2}$$

$$\vdots$$

$$\leq \|\phi\|_{L^2} \prod_{l=1}^{J-k-1} \|a^{(J-l)}\| \bigg( \sum_n \sum_{r_0} \cdots \sum_{r_{J-k-2}} \bigg| \sum_{r_{J-k-1}} \Delta_{r_{J-k-1}} A^{(k),e}_{r_0,\ldots,r_{J-k-1}} \bigg|^2 \bigg)^{1/2}$$

$$\leq \|\phi\|_{L^2} \underbrace{\prod_{l=1}^{J-k-1} \|a^{(J-l)}\|}_{\leq 1} \bigg( \sum_{r_{J-k-1}} |\Delta_{r_{J-k-1}}|^2 \sum_n \sum_{r_0} \cdots$$

$$\times \sum_{r_{J-k-2}} \bigg| A^{(k),e}_{r_0,\ldots,r_{J-k-1}} \bigg|^2 \bigg)^{1/2}$$

$$\leq \|\phi\|_{L^2} \|\Delta\| \|\overline{d}^{(k),e}\|$$

$$\leq \|\phi\|_{L^2} \|\Delta\| 2^{Jd/2} |\langle f, \psi \rangle_{L^2}|$$

again using Proposition 10 in the last inequality.

**Perturbation of** $\beta^{(k),e}$. This is similar to the previous one, with the exception that we have to deal with the nonlinear activation function. Since it is Lipschitz, that is not a problem. Notice that the perturbation $\beta_m^{(k),e} \mapsto \beta_m^{(k),e} + \Delta_m$ induces a change $F(f) \mapsto F(f) + \xi$ for

$$\xi := \sum_n \phi_n(x) \sum_{r_0} a^{(J-1)}_{r_0} \cdots \sum_{r_{J-k-1}} b^{(k),e}_{r_{J-k-1}} D^{(k),e}_{r_0,\ldots,r_{J-k-1}}$$

where

$$D^{(k),e}_{r_0,\ldots,r_{J-k-1}} := \rho(d^{(k),e,\Delta}_{2^{-J+k+1}n+2^{-J+k}r_0\cdots+2^{-1}r_{J-k-1}}) - \rho(d^{(k),e}_{2^{-J+k+1}n+2^{-J+k}r_0\cdots+2^{-1}r_{J-k-1}})$$

and

$$d^{(k),e,\Delta}_n := d^{(k),e}_n + \sum_m \Delta_m s^{(k+1)}_{2n-m}.$$

Accordingly, using the same inequalities as above, we have

$$\|F(f) - F_\Delta(f)\| \leq \|\phi\|_{L^2} \prod_{l=1}^{J-k-1} \|a^{(J-l)}\| \|b^{(k),e}\| \|\rho(d^{(k),e,\Delta}) - \rho(d^{(k),e})\|$$
$$\leq \|\phi\|_{L^2} \|d^{(k),e,\Delta} - d^{(k),e}\| \qquad (26)$$

using the Lipschitz continuity of the activation function $\rho$. Using the definition of $d^{(k),e,\Delta}$ we have

$$\|d^{(k),e,\Delta} - d^{(k),e}\| \leq \|\Delta\| \|s^{(k+1)}\| \leq \|\Delta\| 2^{Jd/2} |\langle f, \psi \rangle_{L^2}|$$

using again the Proposition 10. This completes this part of the claim.

**Perturbation of $\tau_k$.** This is similar to the perturbation of $\beta^{(k),e}$. In fact, notice that the perturbation $\tau_k \mapsto \tau_k + \Delta$ induces a change $F(f) \mapsto F(f) + \xi$ for

$$\xi = \sum_n \phi_n(x) \sum_{r_0} a_{r_0}^{(J-1)} \cdots \sum_{r_{J-k-1}} \sum_e b_{r_{J-k-1}}^{(k),e} E_{r_0,\ldots,r_{J-k-1}}^{(k),e}$$

where

$$E_{r_0,\ldots,r_{J-k-1}}^{(k),e} := \overline{d}_{2^{-J+k+1}n + 2^{-J+k}r_0 \cdots + 2^{-1}r_{J-k-1}}^{(k),e,\Delta} - \overline{d}_{2^{-J+k+1}n + 2^{-J+k}r_0 \cdots + 2^{-1}r_{J-k-1}}^{(k),e}$$

and

$$\overline{d}_n^{(k),e,\Delta} = \sigma(d_n^{(k),e} - \tau_k - \Delta) - \sigma(-d_n^{(k),e} - \tau_k - \Delta).$$

Hence we have

$$|\overline{d}_n^{(k),e,\Delta} - \overline{d}_n^{(k),e}| \leq |\Delta| \quad \text{for any } n,k,e.$$

We can now bound the perturbation as follows

$$\int \left| \sum_n \phi_n(x) \sum_{r_0} a_{r_0}^{(J-1)} \cdots \sum_{r_{J-k-1}} \sum_e b_{r_{J-k-1}}^{(k),e} E_{r_0,\ldots,r_{J-k-1}}^{(k),e} \right|^2 dx$$

$$\leq \|\phi\|_{L^2}^2 \sum_n \left| \sum_{r_0} a_{r_0}^{(J-1)} \cdots \sum_{r_{J-k-1}} \sum_e b_{r_{J-k-1}}^{(k),e} E_{r_0,\ldots,r_{J-k-1}}^{(k),e} \right|^2$$

$$\leq \|\phi\|_{L^2}^2 \sum_n \left( \sum_{r_0} |a_{r_0}^{(J-1)}| \cdots \sum_{r_{J-k-1}} \sum_e |b_{r_{J-k-1}}^{(k),e}| \left| E_{r_0,\ldots,r_{J-k-1}}^{(k),e} \right| \right)^2$$

$$\leq \|\phi\|_{L^2}^2 |\Delta|^2 \, 2^{Jd} \, \|a_{r_0}^{(J-1)}\|^2 \cdots \left( \sum_e \|b_{r_{J-k-1}}^{(k),e}\| \right)^2$$

$$\leq 2^{Jd} 2^{2d} \|\phi\|_{L^2}^2 |\Delta|^2.$$

This yields

$$\|F(f) - F_\Delta(f)\| \leq 2^{Jd/2} 2^d \|\phi\|_{L^2} |\Delta|.$$

**Perturbation of $\alpha^{(k)}$.** This is the most cumbersome argument. We begin by noticing that perturbing $\alpha_m^{(k)}$ to $\alpha_m^{(k)} + \Delta_m$, the SU-net is perturbed to $F(f) \mapsto F(f) + \xi$ where

$$\xi = \sum_n \phi_n(x) \sum_{r_0} a_{r_0}^{(J-1)} \cdots \sum_{r_{J-k}} \sum_e b_{r_{J-k}}^{(k-1),e} G_{r_0,\ldots,r_J}^{(k-1),e}$$

$$+ \sum_n \phi_n(x) \sum_{r_0} a_{r_0}^{(J-1)} \cdots \sum_{r_{J-k+1}} \sum_e b_{r_{J-k+1}}^{(k-2),e} G_{r_0,\ldots,r_J}^{(k-2),e}$$

$$\vdots$$

$$+ \sum_n \phi_n(x) \sum_{r_0} a_{r_0}^{(J-1)} \cdots \sum_{r_{J-1}} \sum_e b_{r_{J-1}}^{(0),e} G_{r_0,\ldots,r_J}^{(0),e}$$

$$+ \sum_n \phi_n(x) \sum_{r_0} a_{r_0}^{(J-1)} \cdots$$

$$\times \sum_{r_{J-1}} a_{r_{J-1}}^{(0)} \left( s_{2^{-J}n+2^{-J}r_0\cdots+2^{-1}r_{J-1}}^{(0),\Delta} - s_{2^{-J}n+2^{-J}r_0\cdots+2^{-1}r_{J-1}}^{(0)} \right)$$

where

$$G_{r_0,\ldots,r_J}^{(l),e} := \rho(d_{2^{-J+l}n+2^{-J+l}r_0\cdots+2^{-1}r_{J-l-1}}^{(l),e,\Delta}) - \rho(d_{2^{-J+l}n+2^{-J+l}r_0\cdots+2^{-1}r_{J-l-1}}^{(l),e})$$

for $l = 0, \ldots, k-1$, and the perturbed $d$ terms are given by

$$d_n^{(k-s),e,\Delta}$$
$$= d_n^{(k-s),e} + \sum_{m_s} \beta_{m_s}^{(k-s),e} \sum_{m_{s-1}} \alpha_{m_{s-1}}^{(k-s+1)} \cdots \sum_{m_1} \alpha_{m_1}^{(k-1)} \sum_{m_0} \Delta_{m_0} s_{2^{s+1}n - 2^s m_s \cdots - m_0}^{(k+1)}$$

for any $s = 1 \ldots, k$, $e = 1, \ldots, 2^d - 1$. And the perturbed $s^{(0)}$ term is given by

$$s_n^{(0),\Delta} = s_n^{(0)} + \sum_{m_k} \alpha_{m_k}^{(0)} \sum_{m_{k-1}} \alpha_{m_{k-1}}^{(1)} \cdots \sum_{m_1} \alpha_{m_1}^{(k-1)} \sum_{m_0} \Delta_{m_0} s_{2^{k+1}n - 2^k m_k \cdots - m_0}^{(k+1)}. \tag{27}$$

Plugging in these expressions above yields

$$\|F(f) - F_\Delta(f)\|_{L^2}$$
$$\leq \|\phi\|_{L^2} \prod_{t=1}^J \|\alpha^{(J-t)}\| \|s^{(0),\Delta} - s^{(0)}\|$$
$$+ \sum_{l=1}^k \|\phi\| \prod_{t=1}^{J-k+l+1} \|\alpha^{(J-t)}\| \sum_e \|\beta^{(k-l),e}\| \|\rho(d^{(k-l),e,\Delta}) - \rho(d^{(k-l),e})\|$$
$$\leq (1 + 2^d k) \|\phi\|_{L^2} \|\Delta\| \|s^{(k+1)}\|$$
$$\leq (1 + 2^d k) \|\phi\|_{L^2} \|\Delta\| \, 2^{Jd/2} |\langle f, \psi \rangle_{L^2}|,$$

which gives the claim.

**Perturbation of $\phi$.** For the statement concerning the perturbation of $\phi$, notice that we have $\phi_k \mapsto \phi_k + \Delta(x) \Rightarrow F(f) \mapsto F(f) + \sum_k \Delta(x-k) \overline{s}_k^{(J)}$, whence

$$\|F(f) - F_\Delta(f)\|_{L^2} \leq \|\Delta\|_{L^2} \|\overline{s}^{(J)}\| \leq (1 + 2^d J) \|\Delta\|_{L^2} \, 2^{Jd/2} |\langle f, \psi \rangle_{L^2}|$$

by Proposition 10

**Perturbation of $\psi$.** The argument here is essentially like for $\alpha^{(k)}$: perturbing $\psi$ to $\psi + \Delta$, the SU-net is perturbed to $F(f) \mapsto F(f) + \xi$ where

$$\xi = \sum_n \phi_n(x) \sum_{r_0} \sum_e b_{r_0}^{(J-1),e} \left( \rho(d_{2^{-1}n+2^{-1}r_0}^{(J-1),e,\Delta}) - \rho(d_{2^{-1}n+2^{-1}r_0}^{(J-1),e}) \right)$$

$$+ \sum_n \phi_n(x) \sum_{r_0} a^{(J-1)}_{r_0}$$
$$\times \sum_{r_1}\sum_e b^{(J-2),e}_{r_1}\left(\rho(d^{(J-2),e,\Delta}_{2^{-2}n+2^{-2}r_0+2^{-1}r_1}) - \rho(d^{(J-2),e}_{2^{-2}n+2^{-2}r_0+2^{-1}r_1})\right)$$

$$\vdots$$

$$+ \sum_n \phi_n(x) \sum_{r_0} a^{(J-1)}_{r_0} \cdots$$
$$\times \sum_{r_{J-1}} \sum_e b^{(0),e}_{r_{J-1}}\left(\rho(d^{(0),e,\Delta}_{2^{-J}n+2^{-J}r_0\cdots+2^{-1}r_{J-1}}) - \rho(d^{(0),e}_{2^{-J}n+2^{-J}r_0\cdots+2^{-1}r_{J-1}})\right)$$
$$+ \sum_n \phi_n(x) \sum_{r_0} a^{(J-1)}_{r_0} \cdots$$
$$\times \sum_{r_{J-1}} a^{(0)}_{r_{J-1}}\left(s^{(0),\Delta}_{2^{-J}n+2^{-J}r_0\cdots+2^{-1}r_{J-1}} - s^{(0)}_{2^{-J}n+2^{-J}r_0\cdots+2^{-1}r_{J-1}}\right)$$

where $d^{(J-s),e,\Delta}$ are defined as

$$d^{(J-s),e,\Delta}_n = d^{(J-s),e}_n + \sum_{m_s} \beta^{(J-s),e}_{m_s} \sum_{m_{s-1}} \alpha^{(J-s+1)}_{m_{s-1}} \cdots \sum_{m_1} \alpha^{(J-1)}_{m_1} \langle \Delta_{2^{s+1}n-2^s m_s\cdots-2m_1}, f\rangle$$

and $s^{(0),\Delta}$ is defined as in (27). Consequently we have

$$\|d^{(J-s),e,\Delta} - d^{(J-s),e}\|^2$$
$$= \sum_k \left| \sum_{m_s} \beta^{(J-s),e}_{m_s} \sum_{m_{s-1}} \alpha^{(J-s+1)}_{m_{s-1}} \cdots \sum_{m_1} \alpha^{(J-1)}_{m_1} \langle \Delta_{2^{s+1}n-2^s m_s\cdots-2m_1}, f\rangle \right|^2$$
$$\leq \underbrace{\|\beta^{(J-s),e}\|^2 \cdots \|\alpha^{(J-1)}\|^2}_{\leq 1} \sum_k \sum_{m_s} \cdots \sum_{m_1} |\langle \Delta_{2^{s+1}n-2^s m_s\cdots-2m_1}, f\rangle|^2$$
$$\leq \sum_{k\in\Gamma_J} |\langle \Delta_k, f\rangle|^2 \leq 2^{Jd}|\langle f, \Delta\rangle_{L^2}|.$$

Using now the same argument as for $\alpha^{(k)}$ completes the claim. □

## Acknowledgments

The author was funded by the Deutsche Forschungsgemeinschaft (DFG; German Research Foundation) Postdoctoral Fellowship AL 2483/1-1. The author also wants to thank Johannes Schmidt-Hieber for his support, many discussions and insightful ideas and suggestions about the contents of this paper. The author also thanks the two anonymous referees for the constructive comments.